\documentclass{ecai/ecai} 
\PassOptionsToPackage{sort}{natbib}

\usepackage{hyperref}
\usepackage{amssymb}
\usepackage{amsmath}
\usepackage{amsthm}
\usepackage{amsfonts}
\usepackage{booktabs}
\usepackage{graphicx}
\usepackage{color}

\usepackage{nicefrac}       %
\usepackage{xcolor}         %
\usepackage{algorithmic}
\usepackage{algorithm}

\usepackage{colonequals}
\usepackage{xspace}
\usepackage{xcolor}

\usepackage{nicefrac}
\usepackage{tikz}
\usepackage{multirow}
\usepackage{url}
\usepackage{todonotes}
\usepackage{mdframed}
\usetikzlibrary{patterns,arrows,arrows.meta,calc,shapes,shadows,decorations.pathmorphing,decorations.pathreplacing,automata,shapes.multipart,positioning,shapes.geometric,fit,circuits,trees,shapes.gates.logic.US,fit}
\usetikzlibrary{backgrounds,scopes}
\usepackage{braket}

\usepackage{silence}
\usepackage[inline,shortlabels]{enumitem}
\setlist*[enumerate]{label=(\arabic*)}

\usepackage{titlecaps}

\usepackage{cleveref}

\newcommand{\prism}{\textrm{PRISM}\xspace}
\newcommand{\storm}{\textrm{Storm}\xspace}

\newcommand{\tuple}[1]{\langle #1 \rangle}

\newcommand{\ie}{\emph{i.e.}\@\xspace}

\newcommand{\eg}{\emph{e.g.}\@\xspace}

\DeclareMathOperator*{\argmax}{argmax}
\DeclareMathOperator*{\argmin}{argmin}

\DeclareMathOperator*{\arginf}{arginf}

\DeclareMathOperator*{\softmax}{softmax}
\DeclareMathOperator{\sign}{sign}

\newtheorem{definition}{Definition}
\newtheorem{lemma}{Lemma}
\newtheorem{theorem}{Theorem}

\newcommand{\R}{\mathbb{R}}

\newcommand{\N}{\mathbb{N}}
\newcommand{\I}{\mathbb{I}}

\newcommand{\cM}{\mathcal{M}}

\newcommand{\cT}{\mathcal{T}}
\newcommand{\cD}{\mathcal{D}}

\newcommand{\cB}{\mathcal{B}}
\newcommand{\cH}{\mathcal{H}}
\newcommand{\cO}{\mathcal{O}}
\newcommand{\cQ}{\mathcal{Q}}
\newcommand{\cV}{\mathcal{V}}

\newcommand{\Distr}{\Delta}
\newcommand{\given}{\,\vert\,}

\newcommand{\memupdate}{\eta}
\newcommand{\actionmap}{\delta}
\newcommand{\fscjointprob}{\psi}

\newcommand{\fsctup}{\braket{ N, n_0, \actionmap, \memupdate }}

\newcommand{\TrsFun}{T}
\newcommand{\RewFun}{C}
\newcommand{\ObsFun}{O}

\newcommand{\beliefs}{\cB}

\newcommand{\uTrsFun}{\cT}
\newcommand{\uObsFun}{\cO}

\newcommand{\Vpes}{\underline{\cV}}

\newcommand{\pirnn}{\pi^{\phi}}

\newcommand{\QMDP}{\ensuremath{Q\textsubscript{MDP}}\@\xspace}
\newcommand{\QFIB}{\ensuremath{Q\textsubscript{FIB}}\@\xspace}

\newcommand{\RNN}{\text{RNN}}

\newcommand{\softminpolicy}{\sigma^\phi}

\newcommand{\fsc}[0]{\pi_{f}}

\newcommand{\bndim}[0]{B_h}
\newcommand{\quantvals}[0]{\beta}

\newcommand{\scp}[0]{\textsc{SCP}\@\xspace}
\newcommand{\SCP}[0]{\scp}
\newcommand{\baseline}[0]{baseline\@\xspace}

\newcommand{\Vsa}[0]{\Vpes^{\fsc}_{\star}}
\newcommand{\Vmc}[0]{\Vpes^{\fsc}}
\newcommand{\Vsta}[0]{\widehat{\Vpes}^{\fsc}}
\newcommand{\Vdyn}[0]{\Vmc}
\newcommand{\staV}{\Vsta}
\newcommand{\dynV}{\Vdyn}
\newcommand{\pickT}{\underline{P}^{\fsc}}

\newcommand{\cumcost}[0]{\rho_{\lozenge G}}
\newcommand{\expcost}[0]{J^{\pi}}

\newcommand{\rfscnet}[0]{\textsc{rFSCNet}\@\xspace}
\newcommand{\ours}[0]{\texorpdfstring{\rfscnet}{rFSCNet}}
\newcommand{\framework}[0]{\texorpdfstring{\textsc{PIP}\@\xspace}{PIP}}

\newcommand{\reals}[0]{\mathbb{R}}
\newcommand{\realsforcosts}[0]{\reals_{\geq 0} \cup \{+\infty\}}
\newcommand{\mc}[0]{\mathcal{M}^{\fsc}}
\newcommand{\robvalfsc}{\mathcal{J}^{\fsc}_{\uTrsFun}}
\newcommand{\adverb}[0]{pessimistic\xspace}
\newcommand{\adverbtitle}[0]{\texorpdfstring{Pessimistic\xspace}{Pessimistic}}

\newcommand{\uTrsFunMC}[0]{\mathcal{P}^{\fsc}}
\newcommand{\TrsFunMC}[0]{P^{\fsc}}

\newcommand{\memsize}{|N|}

\newcommand{\pesspomdp}{\underline{M}}

\newif\ifappendix
\global\appendixtrue

\begin{document}

\begin{frontmatter}

\paperid{5931}

\title{Pessimistic Iterative Planning with 
RNNs\\
for Robust POMDPs}

\author[A]{\fnms{Maris~F.~L.}~\snm{Galesloot}%
\thanks{Corresponding Author. Email: maris.galesloot@ru.nl.
\ifappendix
The code is on Zenodo, accessible from: \url{https://doi.org/10.5281/zenodo.16947259}.
\else
The article with appendix is on \textit{arXiv}~\cite{DBLP:journals/corr/abs-2408-08770}.
Code: \url{https://doi.org/10.5281/zenodo.16947259}.
\fi
}%
}
\author[B]{\fnms{Marnix}~\snm{Suilen}%
}
\author[C]{\fnms{Thiago}~\snm{D.~Sim{\~{a}}o}%
}
\author[D]{\fnms{Steven}~\snm{Carr}%
} 
\author[E]{\fnms{Matthijs~T.~J.}~\snm{Spaan}%
} 
\author[D]{\fnms{Ufuk}~\snm{Topcu}%
} 
\author[F,A]{\fnms{Nils}~\snm{Jansen}%
} 

\address[A]{Radboud University Nijmegen, NL}%
\address[B]{University of Antwerp -- Flanders Make, BE}%
\address[C]{Eindhoven University of Technology, NL}%
\address[D]{University of Texas at Austin, US}%
\address[E]{Delft University of Technology, NL}%
\address[F]{Ruhr-University Bochum, DE}%

\begin{abstract}
        \emph{Robust POMDPs} extend classical POMDPs to incorporate model uncertainty
    using so-called \emph{uncertainty sets} on the transition and observation functions, effectively defining ranges of probabilities.
    Policies for robust POMDPs must be (1) memory-based to account for partial observability and (2) robust against model uncertainty to account for the worst-case probability instances from the uncertainty sets.
    To compute such robust memory-based policies, we propose the \emph{pessimistic iterative planning}~(\framework) framework, which alternates between (1)~selecting \emph{\adverb} POMDPs via worst-case probability instances from the uncertainty sets, and (2)~computing finite-state controllers~(FSCs) for these \adverb POMDPs.
    Within \framework, we propose the \ours algorithm, which optimizes a recurrent neural network to compute the FSCs.
    The empirical evaluation shows that \ours can compute better-performing robust policies than several baselines and a state-of-the-art robust POMDP~solver.

\end{abstract}

\end{frontmatter}

\section{Introduction}\label{sec:introduction}

\emph{Robust partially observable Markov decision processes} (RPOMDPs) extend standard POMDPs with sets of probabilistic transition and observation functions~\cite{DBLP:conf/icml/Osogami15}.
These \emph{uncertainty sets} account for imprecision that may, for instance, come from probabilities derived from data, sensors with limited precision, or domain experts expressing uncertainty~\cite{DBLP:conf/cdc/Kalyanasundaram02,DBLP:books/daglib/0014221}.
Such settings are unsuitable for the standard POMDP assumption that probabilities are precisely known.

\emph{Policies} select actions based on limited state information towards some objective, for instance, optimizing the expected reward.
These policies require \emph{memory} as they inherently depend on the sequences of past actions and observations, also known as the \emph{history}. 

\emph{Robust policies} for RPOMDPs must furthermore account for model uncertainty \emph{pessimistically}, that is, they optimize against the \emph{worst-case} instances within the uncertainty sets, providing a lower bound on their actual performance and ensuring their robustness.
In summary, RPOMDPs require policies that (1) use memory to reason over the histories and (2) account for model uncertainty.

To robustly optimize a policy against pessimistic instances of the RPOMDP, a policy's \emph{worst-case} performance needs to be assessed analytically.
Finite-state controllers~(FSCs) are a suitable policy representation 
for which we can find the worst-case performance analytically through efficient robust dynamic programming~\cite{DBLP:journals/mor/Iyengar05}.
However, computing FSCs often relies on selecting a predetermined memory size and structure~\cite{DBLP:conf/uai/Junges0WQWK018}.
Instead, we desire a more flexible memory structure, which we can achieve by optimizing a \emph{recurrent neural network}~(RNN) to search for FSCs~\cite{DBLP:journals/jair/Carr0T21}.
An RNN has a flexible memory structure that can learn sufficient statistics of the histories from data~\cite{DBLP:journals/tmlr/LambrechtsBE22}.
The drawback, however, is that analytically determining the worst-case performance of an RNN-based policy is intractable.

While optimizing policies for POMDPs is extensively studied~\cite{DBLP:conf/nips/Hansen97,aberdeen2003,poupart2004bounded,kurniawati2008sarsop}, methods to compute robust policies for RPOMDPs are sparse, with the most notable approaches being robust 
point-based value iteration~(RPBVI)~\cite{DBLP:conf/icml/Osogami15} and sequential convex programming (SCP)~\cite{DBLP:conf/aaai/Cubuktepe0JMST21}.
Critically, RPBVI scales poorly to large state spaces, and although SCP scales well, it requires a pre-specified FSC structure, which, as shown in our experimental evaluation, may deteriorate performance.
These drawbacks demonstrate the need for flexible and scalable techniques to compute robust policies for RPOMDPs.

\begin{figure}[t]
     \centering
     \resizebox{0.48\textwidth}{!}{%
     \newcommand{\nodewidth}{2cm}
\newcommand{\nodegaph}{2.2cm}
\newcommand{\nodegapv}{1cm}
\newcommand{\textfill}[0]{white}
	
	\tikzset{item node/.style={rectangle, draw, text width=2.8cm, text badly centered, node distance=\nodewidth, inner sep=5pt,rounded corners=0.1cm,minimum height=1.5cm,minimum width=2.5cm}}
	\tikzset{policy/.style={diamond,draw,text width=1.0cm,text badly centered,minimum height=0.5cm,rounded corners=0.2cm}}
	\tikzset{data/.style={ellipse,draw,text width=1.0cm,text badly centered,minimum height=1.0cm}}
	\tikzset{joincircle/.style={draw,circle,fill=black,inner sep=0pt, outer sep=0pt,minimum width=0.2cm}}

	\begin{tikzpicture}
		\node[item node, draw = white] (model-based-learning) {(1) \textbf{Supervised learning of robust FSCs}};
  
		\node[item node, draw = white, very thick, right=\nodegaph of model-based-learning] (adversarial-model) {(2) \textbf{\titlecap{\adverb} POMDP selection}};
  
		\node[item node, draw = white, below=\nodegapv of adversarial-model, dashed] (evaluation) {Robust policy evaluation};
  
		\node[item node, below=\nodegapv of model-based-learning, draw = white] (outputbox) {RNN $\rightarrow$ FSC};
  
		\draw[draw = black, fill=none, rounded corners=0.1cm] (model-based-learning.north west) rectangle (outputbox.south east);
  
		\draw[very thick] (adversarial-model.west) edge[-latex'] node[fill=\textfill,text width=1.15cm, text centered]{POMDP} (model-based-learning.east);
  
		\draw[very thick] (model-based-learning.south) edge[-latex'] node[fill=\textfill,text width=1.15cm, text centered]{RNN} (outputbox.north);
  
		\draw[very thick] (outputbox.east) edge[-latex'] node[fill=\textfill,text width=0.75cm, text centered]{FSC} (evaluation.west);
  
		\draw[very thick] (evaluation.north)  edge[-latex']node[fill=\textfill,text width=2cm, text centered]{Robust value} (adversarial-model.south);
  
  		\draw[draw = black, fill=none, rounded corners=0.1cm] (adversarial-model.north west) rectangle (evaluation.south east);

        \draw[draw = black, dashed, fill=none, rounded corners=0.1cm, label=RPOMDP] ($(adversarial-model.north west) + (-0.25,0.25)$) rectangle ($(evaluation.south east) + (0.25,-0.25)$);

        \node[draw = none, fill=none, above=0.25 of adversarial-model] {RPOMDP};

	\end{tikzpicture}
     }%
\caption{Overview of the \framework framework.
The steps on the left, generating an FSC for \adverb POMDPs through an RNN, are specific to \ours.
}
\vspace{3em}
\label{fig:highlevel-overview}
\end{figure}

\subsection*{Our Approach and Contributions}
We present a new approach for computing robust memory-based policies for RPOMDPs.
We propose a general framework for planning in RPOMDPs, and by combining the strengths of the flexibility of memory representation of RNNs and the exact robust performance evaluation of FSCs, an algorithmic instance of this framework where we leverage RNNs to compute FSC policies for RPOMDPs.
Specifically, our contributions are:

\medskip

\noindent
    \textbf{Pessimistic iterative planning for RPOMDPs}~(\Cref{sec:method})\textbf{.} %
    We present the \emph{pessimistic iterative planning}~(\framework) framework to find robust policies for RPOMDPs, as outlined in \Cref{fig:highlevel-overview}.
    \framework iteratively computes policies for POMDPs within the uncertainty sets of the RPOMDP that are \emph{\adverb} instances to the current policy, \ie, a worst-case instance given the current policy.
    \framework alternates between two main steps: (1) computing FSCs for (\adverb) POMDPs and (2) evaluating the FSC on the RPOMDP and selecting \adverb POMDPs.
    We implement \framework in \ours, an RNN-based algorithm consisting of two parts corresponding to the steps of \framework:

\begin{enumerate}[leftmargin=*]
    \item \textbf{Supervised learning of robust FSCs}~(\Cref{sec:lhs})\textbf{.} 
    We train an RNN based on data collected from \emph{supervision policies} that we optimize for the \adverb POMDPs.
    From the RNN, we extract the FSC that we use for robust policy evaluation and \adverb POMDP selection. 
    With these \adverb POMDPs, we further train the RNN by refining the collected histories and associated supervision policy,
    guiding the RNN and, therefore, the extracted FSCs towards a robust policy. 
    \item \textbf{\titlecap{\adverb} POMDP selection}~(\Cref{sec:rightside})\textbf{.} 
    First, we compute the worst-case performance of the FSC on the RPOMDP via our robust policy evaluation, thereby producing a guaranteed lower bound on its performance.
    Using the results from the FSC's robust evaluation, we construct a linear program that efficiently finds a POMDP instance within the uncertainty sets that is \adverb, \ie, a worst-case, to the current FSC.
\end{enumerate}
The experimental evaluation on four benchmarks showcases that (1) the FSCs found by \ours are competitive with and in some cases outperform the state-of-the-art SCP solver for RPOMDPs~\cite{DBLP:conf/aaai/Cubuktepe0JMST21}, and (2) using \framework increases robustness compared to baselines that train on POMDPs that are heuristically chosen from the uncertainty sets.

\section{Preliminaries}\label{sec:preliminaries}
The set of all probability distributions over $X$ is denoted by $\Distr(X)$.
A distribution $\mu \in \Distr(X)$ is called \emph{Dirac} if $\mu(x) = 1$ for precisely one $x \in X$ and zero otherwise. 
The number of elements in a set $X$ is denoted by $|X|$.
\emph{Iverson brackets} return $[P] = 1$ if predicate $P$ is true and $0$ otherwise.
Finally, the set of \emph{probability intervals} with lower bounds strictly greater than zero is $\I = \{ [i,j] \mid 0 < i \leq j \leq 1 \}$.

\begin{definition}[POMDP]\label{def:pomdp}
    A \emph{partially observable Markov decision process} is a tuple $M = \braket{S, A, \TrsFun, Z, \ObsFun, \RewFun}$, where $S, A, Z$ are finite sets of states, actions, and observations, $\TrsFun \colon S \times A \to \Distr(S)$ is the transition function, $\ObsFun \colon S  \to Z$ is the deterministic state-based observation function, and $\RewFun \colon S \times A \to \R_{\geq 0}$ is the cost function.
\end{definition}
For simplicity, we consider POMDPs with \emph{deterministic} observations,
which is without loss of generality, as every POMDP can be transformed into one with deterministic observations at the cost of a polynomial increase in the size of the state space~\cite{DBLP:conf/icra/ChatterjeeCGK15}.
    
A \emph{trajectory} in a POMDP is a sequence of states and actions: $\omega_H = s_1a_1s_2\dots s_H \in (S \times A)^{H-1} \times S$, such that $T(s_{t+1} \mid s_t, a_t) > 0$ for $1\leq t < H$.
A \emph{history} is the observable fragment of a trajectory: $h_H = \ObsFun(s_1)a_1\ObsFun(s_2)\dots \ObsFun(s_H) = z_1a_1z_2\dots z_H \in (Z \times A)^{H-1} \times Z$.
The sets of all trajectories and associated histories are $\Omega$ and $\cH$, respectively.
Histories can be summarized into sufficient statistics known as \emph{beliefs}, that is, probability distributions over states~\cite{kaelbling1998planning,astrom1965optimal}.
The set of (reachable) beliefs is $\beliefs \subseteq \Distr(S)$.
The initial state distribution~(belief) is $b_0 \in \beliefs$.
A belief $b\in\beliefs$ can be computed from a history $h\in\cH$ and $b_0$ using Bayes' rule~\cite{Spaan2012}.

A \emph{policy} resolves the action choices in a POMDP and is a function $\pi \colon \cH \to \Distr(A)$ that maps histories to distributions over actions. 
Since beliefs are sufficient statistics for histories, policies may also be belief-based, \ie, of type $\pi \colon \beliefs \to \Distr(A)$.
A policy is deterministic if it only maps to Dirac distributions, and the set of all (history-based) policies is denoted by~$\Pi$.

We focus on
minimizing the expected cost of reaching a given set of goal states $G \subseteq S$, also known as the \emph{stochastic shortest path} (SSP) problem~\cite{DBLP:books/lib/Bertsekas05}.
While we focus the presentation of this paper on the SSP problem, it generalizes to discounted rewards~\cite{DBLP:phd/ndltd/Patek97,DBLP:conf/ijcai/BonetG09}.
For any trajectory~$\omega$, the cumulative cost~$\rho_{\lozenge G} \colon \Omega \to \realsforcosts$~is~\cite{DBLP:conf/sfm/ForejtKNP11}:
\begin{equation*}\label{eq:cumcost}
        \rho_{\lozenge G}(\omega) = \begin{cases}
            \infty & \forall t \in \mathbb{N}, s_t \not\in G,\\
            \sum_{t = 0}^{\min \{ t \; \mid \; s_t \in G \} - 1} C(s_t,a_t) & \text{otherwise.}
        \end{cases}
\end{equation*}
    The objective is to find an optimal policy $\pi\in\Pi$ that minimizes the \emph{expected cost} $\expcost_T$ of 
    policy $\pi$ under transition function $T$:
\begin{equation*}
     \pi^* \in \arginf\limits_{\pi \in \Pi} \expcost_T, \quad \text{where} \quad \expcost_T = \mathbb{E}_{\pi,T} \big[ \cumcost(\omega) \mid s_0 \sim b_0 \big].
\end{equation*}
Here, the expectation $\mathbb{E}_{\pi,T}\left[\cdot\right]$ is over the trajectories $\omega$ generated by following policy $\pi$ under the transition function $T$.
The decision variant of the problem of finding an optimal policy for the SSP problem in a POMDP is undecidable~\cite{DBLP:journals/ai/MadaniHC03}.
Therefore, it is common to approximate optimal policies with finite memory.
A policy is \emph{finite-memory} if it can be represented by a finite-state controller~(FSC)~\cite{DBLP:conf/nips/Hansen97}.
\begin{definition}[FSC]\label{def:fsc}
A \emph{finite-state controller} for a POMDP $M$ is a tuple $\fsc = \fsctup$ where $N$ is a finite set of {memory nodes}, $n_0 \in N$ the {initial node},
$\actionmap \colon N \times Z \to \Distr(A)$ is the {action function}, and $\memupdate \colon N \times Z \to N$ is the {memory update}.
\end{definition}
$\Pi_f \subseteq \Pi$ denotes the set of FSCs.
At execution time, at state $s$ and node $n$, using $z=\ObsFun(s)$, the FSC selects action $a \sim \actionmap(\cdot \given n, z)$ and updates its node to $n' = \memupdate(n, z)$.
By $\fscjointprob(a, n' \mid s, n) = \actionmap(a \given n, \ObsFun(s)) [n'=\memupdate(n, \ObsFun(s))]$, we denote the joint probability of the FSC selecting action $a$ and updating to the next memory node $n'$.
The expected costs $J^{\fsc}_T$ of an FSC $\fsc$ on a POMDP $M$ is evaluated by computing the state-values 
on the product Markov chain~\cite{DBLP:conf/uai/MeuleauKKC99,BK08}.

\section{Robust POMDPs}\label{sec:robust}
Robust POMDPs (RPOMDPs)~\cite{DBLP:conf/icml/Osogami15,ijcai2024p740} extend POMDPs by accounting for uncertainty in the transition and observation functions. 
That is, the probabilities are no longer given, but only known to belong to some \emph{uncertainty set}~\cite{DBLP:journals/mor/WiesemannKR13,DBLP:conf/icml/Osogami15}.
Without loss of generality, we focus on uncertainty over the transition function, as any RPOMDP with an uncertain observation function can be transformed into an equivalent RPOMDP with a deterministic observation function, see 
\ifappendix
\Cref{app:detobs}
\else
Appendix~B~\cite{DBLP:journals/corr/abs-2408-08770}
\fi
or \citep{ijcai2024p740}.
Similarly, we omit reward uncertainty for brevity, but it can be included in a straightforward manner~\cite{DBLP:conf/icml/Osogami15}. 

\begin{definition}[RPOMDP]
    A robust POMDP with interval uncertainty is a tuple $\cM = \braket{S, A, \uTrsFun, \RewFun, Z, \ObsFun}$, where $S, A, Z$, 
    and 
    $\RewFun, \ObsFun$ are 
    as in~\Cref{def:pomdp}, and $\uTrsFun \colon S \times A \to (S \to \I \cup \{0\})$ is the \emph{uncertain transition function} that maps transitions to either a probability interval in $\I$, or probability~$0$ whenever the transition does not exist.
\end{definition}
Henceforth, we assume the standard rules for interval arithmetic~\cite{DBLP:journals/jacm/HickeyJE01}.
We allow only intervals with a lower bound greater than zero to avoid the vanishing of transitions, also known as \emph{graph preservation}. 
This assumption is standard if intervals are learned from data~\cite{DBLP:journals/mor/WiesemannKR13}. 
Furthermore, for the objective considered here, a lower bound of zero is only possible if one assumes all trajectories reach the goal states $G$ with probability one.
Otherwise, $G$ may become unreachable in the worst case.

A robust MDP~(RMDP)~\cite{DBLP:conf/birthday/SuilenBB0025} is a fully observable RPOMDP, \ie, where $Z\equiv S$ and $\forall s\in S\colon O(s) = s$.
A POMDP $M$ with transition function $T$ is called an \emph{instance} of the RPOMDP if every transition probability of $T$ lies within its respective interval in $\cT$.
With abuse of notation, we may also write $T\in\cT$ and $M \in \cM$.

RPOMDPs can be seen as a game between the agent and a second player, \emph{nature}, which selects probability distributions from the uncertainty set.
We assume the \emph{dynamic} uncertainty model~\cite{DBLP:journals/mor/Iyengar05}, which means that nature's choices are not restricted by any previous choice,
and independence between all state-action pairs, a common assumption known as \emph{$(s,a)$-rectangularity}~\cite{DBLP:journals/mor/WiesemannKR13}
Specifically, the uncertain transition function $\uTrsFun$ factorizes over state-action-pairs, \ie, it is comprised of the Cartesian product
$
    \uTrsFun = \bigotimes_{(s,a) \in S \times A} \uTrsFun(s,a),
$
where:
\[
    \uTrsFun(s,a) = \left\{T(s,a) \in\Delta(S) \given \forall s'\in S \colon T(s'\mid s,a) \in \uTrsFun(s,a)(s')\right\}.
\]
While we focus on uncertainty in the form of intervals, the results presented in this paper generalize to $(s,a)$-rectangular uncertainty sets that are graph preserving and form \emph{convex polytopes}, such as those constructed from the $\ell_1$ or $\ell_\infty$ norms.

RPOMDPs have two optimal value functions and associated optimal policies: one where the agent and nature play \emph{adversarially}, and one where they play cooperatively.
The former is the robust (or pessimistic) setting, and the latter is the optimistic setting.
Our approach extends to both the robust and the optimistic case.
We focus the presentation on the robust setting in the remainder of this paper.

In RPOMDPs, the trajectories generated by a policy $\pi\in\Pi$ depend on the transition function $T\in\uTrsFun$.
Hence, we minimize the \emph{robust value}~$\mathcal{J}^{\pi}_{\mathcal{T}}$, which represents the \emph{worst-case expected cost}:
\begin{equation*}
     \pi^* \in \arginf\limits_{\pi \in \Pi} \mathcal{J}^{\pi}_{\uTrsFun}, \quad \text{where} \quad \mathcal{J}^{\pi}_{\uTrsFun} = \sup\limits_{T\in\uTrsFun} \expcost_T.
\end{equation*}

The undecidability of the decision variant of this problem follows from that of our objective in POMDPs~\cite{DBLP:journals/ai/MadaniHC03}.
Therefore, we cannot aim for completeness. 
Instead, we focus on developing a practical and modular algorithm that computes robust FSCs, including a sound policy evaluation, while allowing for a flexible memory structure.

\paragraph{Goal.}
Given an RPOMDP $\cM = \braket{S, A, \uTrsFun, \RewFun, Z, \ObsFun}$,
compute an FSC $\fsc \in \Pi_f$
that minimizes the robust value $\mathcal{J}^{\fsc}_{\mathcal{T}}$.

\section{Pessimistic Iterative Planning}\label{sec:method}
We present our main contributions.
First, we outline the two main parts of the \emph{pessimistic iterative planning}~(\framework) framework.
Subsequently, we give an overview of our algorithmic implementation of \framework, named \ours, which computes robust FSCs for RPOMDPs by optimizing an RNN and extracting FSCs from the RNN.

\subsection{The \framework Framework}
\label{sec:pip}
Analogously to a sequential two-player game, \framework iteratively executes two parts, representing the two sides of \Cref{fig:highlevel-overview}: 
\begin{enumerate}[(1)]
    \item Compute an FSC policy $\fsc$ for a given POMDP $M\in\cM$. 
    \item Select a \emph{\adverb} POMDP $\pesspomdp\in\cM$ with respect to $\fsc$, and set $M\gets \pesspomdp$. 
    Give $M$ as input to (1).
\end{enumerate}
These steps are repeated in a sequential game-like fashion.
The first player executes Step (1) and computes an FSC $\fsc$ that minimizes the expected cost in the (non-robust) POMDP $M$.
Note that \framework may use any existing approach that computes FSCs for (non-robust) POMDPs.
The other player then executes Step (2) and determines a POMDP $\pesspomdp\in\mathcal{M}$ that is pessimistic with respect to $\fsc$, effectively \emph{maximizing} the expected cost incurred under $\fsc$.
Returning to Step (1), the first player takes $\pesspomdp$ into account and computes a new (updated) FSC optimized for $\pesspomdp$.
This process is repeated until we reach a \emph{termination criterion}, \ie, the robust performance of the FSC satisfies a target threshold or a maximum number of iterations is reached.

\paragraph{On convergence.}
Computing robust policies for cost minimization in an RPOMDP results in a sequential zero-sum game between the agent and nature~\cite{DBLP:conf/ijcai/HorakBC18,ijcai2024p740}.
PIP approximates this game through an iterated best-response formulation, where both the agent and nature optimize their policy based on the given policy of the other player. 
The goal of this paper is not to formalize the resulting game.
Still, we present our considerations in the following.
In finite games where each player optimizes over a finite set of policies, convergence is achieved when the optimization of both players' policies reaches a saddle point, \ie, $\inf_{\fsc \in \Pi_f} \sup_{T\in\mathcal{T}} J_{T}^{\fsc} = \sup_{T\in\mathcal{T}} \inf_{\fsc \in \Pi_f} J^{\fsc}_{T}$, 
and a Nash equilibrium follows under mild conditions~\cite{peters2015game}.
If we limit the agent to deterministic FSCs, it can be argued that the set of policies of both players is finite.
However, the sequential structure of the PIP approximation does not fit directly into the classical Nash formulation.
Stackelberg games encapsulate sequential games~\cite{DBLP:journals/automatica/ZhengJL22,peters2015game}, but assume that the first player considers all possible responses of the other player, which does not occur in PIP.
Establishing whether equilibria exist and whether a formal guarantee of convergence for a framework like PIP is possible is an open problem. 
Therefore, PIP does not aim at completeness; we cannot guarantee that an optimal policy or a set threshold will be achieved in the long run of PIP. 
Instead, the PIP framework provides soundness: the robust policy evaluation steps provide a bound on the worst-case performance.
Together with a preset threshold, this evaluation provides a sound termination criterion and an additional limit to the number of iterations.

\subsection{The \ours Algorithm}
Next, we detail the steps of our algorithm \ours, which implements the two parts of~\framework.
Part one corresponds with \Cref{sec:lhs} and the left-hand side in \Cref{fig:highlevel-overview}.
We compute FSCs for input POMDPs~$M\in\cM$ using an RNN, which is specific to \ours:
\begin{enumerate}[i]
    \item Compute a \emph{supervision policy} $\pi_M$ for the input POMDP~$M$~(\Cref{subsec:supervision:policies}) and simulate $\pi_M$ on $M$ to collect the histories and the action distributions of $\pi_M$ into a data set $\cD$
    (\Cref{subsec:data:generation}).
    \item Train the RNN policy $\pirnn$ on the data set $\cD$~(\Cref{subsec:rnn:training}) and extract an FSC $\fsc$~(\Cref{subsec:extract:fsc}).
\end{enumerate}
Part two corresponds to \Cref{sec:rightside} and the right-hand side of \Cref{fig:highlevel-overview}.
Here, we implement the robust policy evaluation of a computed FSC and, subsequently, the selection of new \adverb POMDPs.
These steps 
other ways of computing the FSC.
In this part, we:
\begin{enumerate}[i]
    \setcounter{enumi}{2}
    \item Evaluate the robust value $\robvalfsc$ of the FSC~$\fsc$ through robust dynamic programming (\Cref{subsec:robust_eval}).
    \item %
    Compute a new \adverb POMDP $\pesspomdp \in \cM$ based on the FSC~$\fsc$~(\Cref{subsec:lp}). Set $M\gets \pesspomdp$ as the next input.
\end{enumerate}
We track the best policy found based on $\robvalfsc$, and we determine whether to stop the algorithm based on the aforementioned termination criteria of the \framework framework.
Otherwise, after Step (iv), we start a new iteration at Step~(i) using $\pesspomdp$ as the new input POMDP.

We opt for a model-based approach, using approximate solvers to compute \emph{supervision policies} $\pi_M$ for $M\in\cM$ and train the RNN to imitate $\pi_M$ in a supervised manner, resembling \emph{imitation learning}~\cite{DBLP:journals/jmlr/RossGB11,DBLP:journals/csur/HusseinGEJ17}.
Alternatively, we could employ model-free techniques from reinforcement learning, such as \emph{recurrent policy gradients}~\cite{DBLP:conf/icann/WierstraFPS07}, to optimize the RNN.
Our method readily supports this.
However, model-free methods neglect the available information from the model.
Therefore, in addition to the fact that reinforcement learning under function approximation may diverge~\cite{DBLP:conf/ijcnn/FairbankA12b}, it typically requires many samples to achieve reasonable values~\cite{DBLP:conf/iros/PetersS06,DBLP:journals/corr/abs-2006-16712}.

\section{Supervised Learning of Robust FSCs}\label{sec:lhs}
In this section, we specify the methods in \ours used to compute FSCs given input POMDPs: collecting data, RNN architecture and training, and extracting a finite-state controller policy from an RNN.

\subsection{Supervision Policies}\label{subsec:supervision:policies}
Since computing an optimal policy in POMDPs for our objective is undecidable (recall \Cref{sec:preliminaries}) and our framework relies on computing policies for multiple POMDPs, fast approximate methods are needed.
We compute \emph{supervision policies} $\pi_M \colon \beliefs \to \Distr(A)$ that approximate the optimal policy $\pi^*$ for the POMDP $M$.
In particular, we approximate the belief-based action-values $Q_M \colon \beliefs \times A \rightarrow \mathbb{R}$ for $M\in\mathcal{M}$, which we denote by $\hat{Q}_M$, and use either \QMDP~\cite{Littman1995} or the fast-informed bound (FIB)~\cite{hauskrecht2000value} to compute these approximations, denoted \QMDP and \QFIB respectively.
The supervision policy is then derived by acting greedily, \ie, taking a Dirac distribution on the minimizing action, such that $\pi_M(a \mid b) = [a = \argmin_{a\in A} \hat{Q}_M(b, a)]$.

\QMDP neglects information-gathering and assumes full state observability after a single step~\cite{Littman1995}.
The values \QFIB are tighter than those of \QMDP since it factors in the observation of the next state. 
See \ifappendix
\Cref{app:supervision_policies}
\else
Appendix~C~\cite{DBLP:journals/corr/abs-2408-08770}
\fi 
for more details.
Other methods may also be used, such as variants of \emph{partially observable Monte Carlo planning}~\cite{silver2010monte} or \emph{heuristic-search value iteration}~\cite{DBLP:conf/uai/SmithS04} for POMDPs with target states.

\subsection{Data Generation}\label{subsec:data:generation}
To train the RNN, we generate a dataset~$\cD$ by simulating the supervision policy $\pi_M$ on the current (\adverb) POMDP $M \in \cM$.
We execute $I \in \N$ simulations up to length $H \in \N$.
During the simulations, we play the actions of the supervision policy $\pi_M$.
We aggregate the histories and associated action distributions generated by following $\pi_M$.

To elaborate, at each time step~$1\leq t < H$ of simulation $i$, we track the beliefs~$b_t^{(i)}$ associated with history~$h^{(i)}_t$, using the transition function $T$ of $M$, with $b_0^{(i)}$ the initial belief $b_0$ of the RPOMDP~$\cM$. 
Then, $\mu_t^{(i)} = \pi_M(\cdot \mid b_t^{(i)})$ is the action distribution of the supervision policy $\pi_M$ during the simulation, and $a^{(i)}_t \sim \mu^{(i)}_t$ is the action used in simulation $i$ at time $t$.
We store the histories and action distributions in the dataset $\cD = \{ \{h_t^{(i)}, \mu_{t}^{(i)}\}^H_{t=1} \}_{i = 1}^{I}$, which then consists of $I\cdot H$ histories and the associated action distributions.
We only consider the data generated by simulation in the most recent iteration, which we found leads to the most stable results.
In the next step, we employ $\cD$ to train the RNN policy $\pirnn$.

\subsection{RNN Policy Architecture and Training}\label{subsec:rnn:training}
Comparably to FSCs, RNNs are (infinite) state machines parameterized by differentiable functions. 
The states $\hat{h}\in\hat{\cH} \subseteq \R^d$ represent memory, where $d$ defines the \emph{hidden size} of the vector.
Analogously to an FSC, we design the RNN to consist of a parameterized internal memory update $\hat{\memupdate}_\phi \colon \hat{\cH} \times Z \to \hat{\cH}$ that recurrently computes the new latent state $\hat{h}\in\hat{\cH}$ from the observations of a history $h\in\cH$.
Thus, the RNN represents a function $\RNN \colon\cH \to \hat{\cH}$.
When we append a fully connected layer with a $\softmax$ activation function $\softminpolicy \colon \hat{\cH} \to \Distr(A)$ to the RNN, it yields an RNN policy network $\pirnn \colon \cH \to \Distr(A)$ from histories to distributions over actions.

The training objective for the RNN is to minimize the distance between the distributions of the RNN policy $\pirnn$ and the distributions~$\mu$ of the supervision policy~$\pi_M$ over the dataset $\cD$: %
$
    \min_\phi \frac{1}{|\cD|H} \sum^{|\cD|}_{i=1} \sum^H_{t=1} 
    L
    \big( \pirnn(h_t^{(i)}) \parallel \mu_t^{(i)} \big),%
$
where $L$ is a distance or loss function, \eg, Kullback-Leibler divergence, and for each batch index $i$ and time-step $t$, the histories~$h_t^{(i)}$ are the RNN's inputs, and the action distributions $\mu_t^{(i)}$ are the labels.
To optimize the parameters $\phi$ of the RNN, we calculate the gradient via backpropagation through time~\cite{DBLP:journals/pieee/Werbos90}.
See 
\ifappendix
\Cref{app:rnn}
\else
Appendix~D.2~\cite{DBLP:journals/corr/abs-2408-08770}
\fi
for more details.

\subsection{Extracting an FSC from an RNN}\label{subsec:extract:fsc}
Recall that in our approach, we change the POMDP at each iteration to be \adverb against the current policy.
For robust policy evaluation and to select the new \adverb POMDP~$\pesspomdp\in\cM$, we need a finite-memory representation of the policy, which we find as follows.

We cluster the hidden memory states of the RNN~\cite {DBLP:journals/neco/ZengGS93,DBLP:journals/nn/OmlinG96} to find 
an FSC. 
Prior work~\cite{DBLP:journals/jair/Carr0T21} uses a quantized bottleneck network (QBN)~\cite{DBLP:conf/iclr/KoulFG19} to reduce the possible memory states to a finite set. 
They train the QBN \emph{post hoc} by minimizing the mean-squared error on the hidden states generated by the RNN.
Alternatively, we can train it \emph{end-to-end} by updating its parameters with the loss from \Cref{subsec:rnn:training}, which we name QRNN for quantized~RNN. %
Moreover, similar to post hoc training of the QBN, we can run a clustering algorithm such as \texttt{k-means++}~\cite{DBLP:conf/soda/ArthurV07} to minimize the in-cluster variance of the hidden states.
For post hoc training, we employ the histories in $\cD$ to generate the RNN hidden states. 
We consider all three methods in our experiments and provide more details of the QBN in 
\ifappendix
\Cref{app:qbn}.
\else
Appendix~D.1~\cite{DBLP:journals/corr/abs-2408-08770}.
\fi

Instead of through simulations, as done by~\citet{DBLP:conf/iclr/KoulFG19}, we utilize the model to construct the FSC.
The clustering determines a finite set $N$ of memory nodes of the RNN's possible hidden states, \ie, a partitioning of $\hat{\cH}$.
We find the FSC's memory update $\memupdate$ by executing a forward pass of the RNN's memory update $\hat{\memupdate}_\phi$ for each reconstruction of ${n} \in {N}$, which produces the next memory nodes ${n}'\in N$ and RNN hidden state $\hat{h}\in\cH$ for each $z\in Z$ by exploiting the batch dimension. 
Then, the action mapping $\actionmap$ for ${n}$ and $z$ is given by the distribution of the RNN policy network $\sigma^\phi(\hat{h})$ for the next memory state.
The RNN's initial state determines the initial node $n_0$, and we prune any unreachable nodes from the FSC.

Alternatives include unfolding $\pirnn$ and minimizing the resulting policy tree~\cite{DBLP:journals/tcyb/GrzesPYH15}, or using active automata learning~\cite{aalrnn}.
Yet, these methods do not adequately scale since the size of the policy tree is exponential in the horizon, which is an inherent problem of POMDPs.

\section{Robust Policy Evaluation and \adverbtitle  Selection of POMDPs}\label{sec:rightside}
The previous section explains how we find FSCs given the input \adverb POMDPs.
Now, we present our methods for a sound robust policy evaluation of the FSC and, subsequently, for selecting a \adverb, \ie, a worst-case POMDP $\pesspomdp \in \cM$ from this evaluation.

\subsection{Robust Policy Evaluation}\label{subsec:robust_eval}
We evaluate the robust performance of the FSC on the RPOMDP via the product construction of a \emph{robust Markov chain}~(RMC), similar to the one used for evaluating FSCs in (non-robust) POMDPs~\cite{DBLP:conf/uai/MeuleauKKC99}.

\begin{definition}[RMC]
\label{def:robust:markov:chain}
From an RPOMDP $\cM = \braket{S,A,\uTrsFun, \RewFun, Z, \ObsFun}$ with initial belief $b_0$ and an FSC $\fsc = \fsctup$ we construct a \emph{robust Markov chain} $\mc = \braket{S \times N, b^{\fsc}_0,\uTrsFunMC, \RewFun^{\fsc}}$ where the state-space is the product of RPOMDP states~$S$ and FSC memory nodes~$N$, the initial state distribution is $b^{\fsc}_0(\braket{s,n}) = b_0(s)[n = n_0]$. 
The uncertain transition and cost functions are as follows:
\begin{align*}
\uTrsFunMC(\braket{s',n'} \given \braket{s, n}) &=
 \sum\limits_{a \in A}\uTrsFun(s, a)(s') 
 \fscjointprob(a, n' \mid s, n),
 \\
 \RewFun^{\fsc}(\braket{s, n}) &=
 \sum\limits_{a \in A} \actionmap(a \given n, \ObsFun(s))\RewFun(s,a).
\end{align*}
\end{definition}

The \emph{robust value} $\robvalfsc$ of the FSC~$\fsc$ is determined by computing the robust state-value $\Vpes^{\fsc} \colon S\times N\to \mathbb{R}_{\geq 0}$ of the RMC~$\mc$ via dynamic programming~\cite{patek_stochastic_1999,DBLP:journals/mor/Iyengar05}, such that its value $\robvalfsc = \sum_{s, n} b_0^{\fsc}(\braket{s,n}) \Vpes^{\fsc}(\braket{s,n})$ is the worst-case expected cost of the FSC given the assumptions of this paper.
The robust state-value $\Vmc$ is the fixed point~\cite{patek_stochastic_1999} of the following robust Bellman equation:
\begin{align}\label{eq:robustv}
    \Vpes^{\fsc}(\braket{s,n}) &= 
    C^{\fsc}(\braket{s,n}) + 
    \sup_{\TrsFunMC(\braket{s,n})\in\uTrsFunMC(\braket{s,n})} \\
    &\sum_{s' \in S}\sum_{n' \in N} \TrsFunMC(\braket{s',n'} \given \braket{s, n}) \Vpes^{\fsc}(\braket{s',n'})
    .\label{eq:robust:as:value}\nonumber
\end{align}
where $\uTrsFunMC(\braket{s,n}) = \sum_{a\in A} \uTrsFun(s,a) \fscjointprob(a, \cdot \mid s, n)$ is the resulting uncertainty set in the robust Markov chain.
If the FSC $\fsc$ reaches the target set $G$ with probability one, then the above Bellman equation is a contraction with respect to a weighted maximum norm~\cite{DBLP:journals/mor/BertsekasT91,DBLP:phd/ndltd/Patek97,patek_stochastic_1999}.

Effectively, $(s,a)$-rectangularity carries over to the product state-space of the RMC, leading to $(\braket{s,n}, a)$-rectangularity; the uncertainty set additionally factorizes over the memory nodes.
That is, nature's choices can change depending on the agent's memory node, yielding a conservative bound on the robust value, as formalized in the following theorem.
\begin{theorem}
    Given the FSC~$\fsc$, the robust state-values~$\Vmc$ of the RMC~$\mc$ provide a conservative (upper) bound on the value~$\Vsa$ of the FSC~$\fsc$ under $(s,a)$-rectangularity in the RPOMDP.
\end{theorem}
Since the agent policy $\fsc$ is fixed, dynamic programming on the RMC is tractable due to both the rectangularity and convexity of the uncertainty sets, and the dynamic uncertainty model~\cite{DBLP:journals/ior/NilimG05,DBLP:journals/mor/Iyengar05,DBLP:journals/ipl/ChenHK13}.

We could consider the $(\braket{s,n}, a)$-rectangularity simply as $(s,a)$-rectangularity plus the dynamic uncertainty model, allowing nature to choose different probabilities when revisiting $(s,a) \in S\times A$ for different $n\in N$.
However, the dynamic model in this case does not converge to a fixed probability distribution, \ie, a static model, for each $(s,a)$ pair as is the case for RMDPs with stationary policies~\cite{DBLP:journals/mor/Iyengar05}.
Instead, it converges to a fixed probability distribution for each $(s,n,a)$ triplet.
It follows from the fact that a static model results in the same value as a dynamic uncertainty model over the state space of the robust Markov chain~\cite{DBLP:journals/mor/Iyengar05,DBLP:journals/ipl/ChenHK13}. 
The intuition here is that the FSC $\fsc$ is a stationary policy over $S\times N$.
Therefore, it also suffices for nature to act with a stationary policy.
The next theorem formalizes this result.
\begin{theorem}
    In the RMC $\mc$, the value under the dynamic model $\dynV$ coincides with the value under a static model $\staV$.
\end{theorem}
\ifappendix
\Cref{appx:rdp} 
\else
Appendix~A~\cite{DBLP:journals/corr/abs-2408-08770}
\fi
contains the proofs for the two theorems above.
Thus, there exists a transition function $T_N \colon S \times N \times A\to \Distr(S)$ that is adversarially chosen by nature that induces the worst-case expected costs of the FSC $\fsc$.
In addition to the usual transition function of a POMDP, it also depends on the memory update function of the FSC.

\subsection{Selecting \adverbtitle  POMDP Instances}\label{subsec:lp}
We now construct a heuristic to find a new POMDP instance $\pesspomdp \in \cM$ that constitutes a local worst-case instance for the current policy $\fsc\in\Pi_f$ under $(s,a)$-rectangularity of the RPOMDP.

Let $\fsc = \braket{N,n_0,\actionmap, \memupdate}$ be the current FSC, and let $\TrsFunMC_T \in \uTrsFunMC$ denote the transition function of the Markov chain by selecting the transition probabilities $T\in \mathcal{T}$.
Given the robust value function $\Vpes^{\fsc}$ computed from \Cref{eq:robustv}, we aim to find a POMDP $\pesspomdp \in \cM$ that induces its worst-case value and, thus, is \adverb to $\fsc$. 
\begin{definition}[\titlecap{\adverb} POMDP]
    Given an FSC policy $\fsc$ and its robust value function~$\Vpes^{\fsc}$, a \emph{\adverb POMDP} $\pesspomdp\in\cM$ is a POMDP $\pesspomdp = \braket{S, A,\underline{\TrsFun}, \RewFun, Z, \ObsFun}$ with a \emph{pessimistic transition function} $\underline{T}\in\mathcal{T}$ with respect to the robust value function~$\Vpes^{\fsc}$, such that
    $
    \underline{T} \in \argmax_{T\in\uTrsFun} \TrsFunMC_T \Vpes^{\fsc}.
    $
\end{definition}
As described in the previous subsection and in more detail in 
\ifappendix
\Cref{appx:rdp}
\else
Appendix A~\cite{DBLP:journals/corr/abs-2408-08770}\fi, the robust value function is effectively computed under $(\braket{s,n},a)$-rectangularity, resulting in probabilities that may differ for each memory node $n$, i.e, with a transition function of type $T_N \colon S\times N \times A \to\Distr(S)$.  
However, we desire our \adverb POMDPs to have a typical transition function of the signature $\underline{T} \colon S \times A \to \Distr(S)$.
Therefore, we compute transition probabilities under the additional constraint that probabilities are independent of the memory nodes $n\in N$, \ie, under $(s,a)$-rectangularity:
\begin{equation*}\label{eq:worst_transfunc}
    \argmax_{{T}(s,a)\in\mathcal{T}(s,a)} 
    \sum_{n, s', n'}
    T(s'\mid s,a) \fscjointprob(a, n' \mid s,n)
    \Vpes^{\fsc}(\braket{s',n'}).
\end{equation*}

We construct a single linear program (LP) that precisely encodes our requirements, using the independence among state-action pairs.
Let $\hat{\TrsFun}_{s,a,s'}$ be the optimization variables representing the probabilities of a pessimistic transition function.
The LP is formulated as:
\begin{align}\label{eq:lp}
\max_{
\hat{\TrsFun}
} 
&\sum\limits_{s,n,a,s',n'}
\hat{T}_{s,a,s'}\;
\fscjointprob(a, n' \mid n, s)\;
\Vpes^{{\fsc}} (\braket{s',n'}) \notag\\
\text{s.t.} \; &\forall s,a \in S\times A\colon\sum\limits_{s'\in S} \hat{\TrsFun}_{s,a,s'} = 1, \\
&\forall s,a,s' \in S\times A\times S\colon\; \hat{\TrsFun}_{s,a,s'} \in \uTrsFun(s,a)(s'). \notag
\end{align}
Solving this LP yields assignments for the variables $\hat{\TrsFun}_{s,a,s'}$ that determine a \emph{pessimistic transition function} $\hat{\underline{\TrsFun}} \colon S\times A \to \Distr(S)$ that satisfies $\hat{\underline{\TrsFun}}(s,a) \in \uTrsFun(s,a)$ for all $(s,a) \in S\times A$.
By construction, the assignments are valid probability distributions within each respective interval and yield a heuristic for the worst possible value for the given FSC under $(s,a)$-rectangularity.

With the selection of a pessimistic POMDP instance $\pesspomdp = \braket{S, A,\hat{\underline{\TrsFun}}, \RewFun, Z, \ObsFun}$, we have closed the loop of \framework, and resume the algorithm by optimizing the FSC for the next input $\pesspomdp\in\mathcal{M}$ unless we have reached a termination criterion specified in \Cref{sec:pip}.

\begin{figure*}[tbp]
    \centering
    \renewcommand\sffamily{}
    \includegraphics[width=\textwidth]{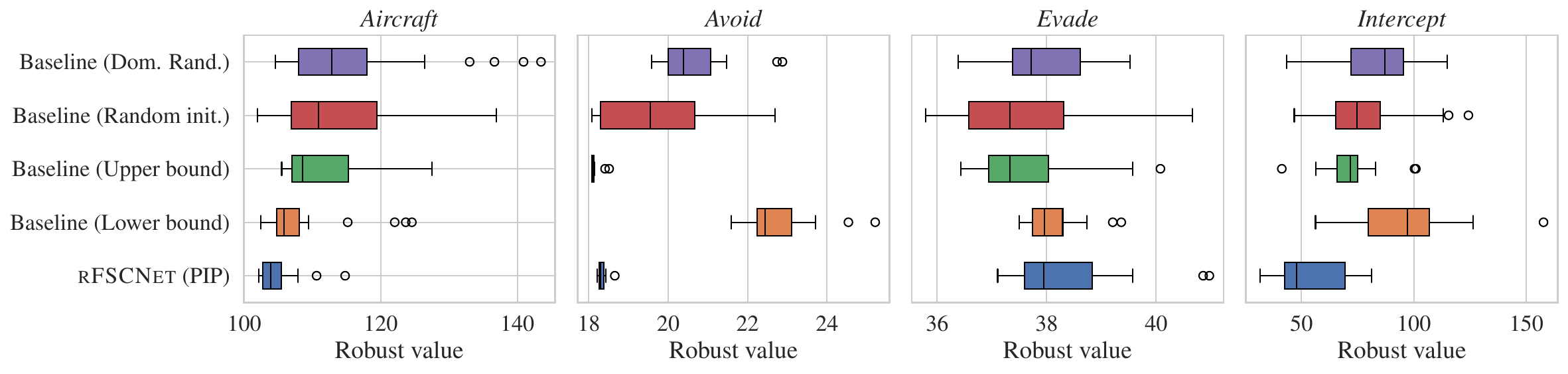}
    \caption{
    Boxplots depicting robust values (lower is better) of the extracted FSC policies for both \rfscnet and the baselines reported across $20$ seeds. 
    All are configured with \QMDP and \texttt{k-means} with $\memsize\leq9$.
    Generally, \ours finds the most robust policies across environments, in contrast to the algorithmically similar baselines, i.e., they employ an RNN to compute FSCs, but do not employ the pessimistic POMDP selection steps of the \framework framework.
    }
    \label{fig:robust_evaluation}
    \vspace{3em}
\end{figure*}

\section{Experimental Evaluation}
\label{sec:experimental:evcaluation}

We empirically assess different aspects of \rfscnet to address the following questions:
\begin{enumerate}[label=\textbf{(Q\arabic*)},leftmargin=*]
    \item \label{q:robustness} \textbf{Robustness and baseline comparison.} 
    How does \rfscnet compare to various baselines that do not utilize the PIP framework: does PIP enable robust performance?
    \item \label{q:comparison} \textbf{Comparison with the state-of-the-art.} 
    How does \rfscnet's performance compare to the \SCP solver?
    \item \label{q:memory} \textbf{Memory size sensitivity.} 
    How does the specified memory size affect \rfscnet and \SCP's performance?
    \item \label{q:configuration} \textbf{Configuration sensitivity.} 
    How do different configurations of supervision policies and clustering methods affect the performance of \rfscnet? 
\end{enumerate}
The experimental evaluation is set up as follows.

\paragraph{Environments.}
We use the existing RPOMDP benchmark of the \emph{Aircraft} collision avoidance problem~\cite{collision,DBLP:conf/aaai/Cubuktepe0JMST21}, and extend three POMDP grid-worlds with adversaries to RPOMDPs, named \emph{Avoid}, \emph{Evade}, and \emph{Intercept}~\cite{DBLP:conf/cav/JungesJS20}.
On \emph{Aircraft}, the agent is tasked to avoid a collision with an intruder aircraft while accounting for uncertainty in the probabilities of the pilot's responsiveness and the intruder changing direction, both mapping to a $[0.6, 0.8]$ interval. 
The grid-world environments model the probability of taking multiple steps instead of a single one for each possible moving action, given by the interval $[0.1, 0.4]$.
We report dimensions in~\Cref{tab:sizes} and
provide 
environment descriptions and run times in
\ifappendix~\Cref{app:benchmark_desc,app:runtimes}
\else
Appendices E and G~\cite{DBLP:journals/corr/abs-2408-08770}\fi, respectively.

\paragraph{Baselines.} 
We evaluate the impact of the \adverb selection of POMDPs of the PIP framework on robust values against several baselines.
These run, like \rfscnet, on POMDPs within the uncertainty sets of the RPOMDP, but selected by one of the following heuristics:
\begin{description}
    \item \textbf{Lower bound.} At the first iteration, select a POMDP within the RPOMDP that greedily assigns the lower bounds of the intervals to transitions, and distribute the remaining probability mass to ensure a valid probability distribution.
    Remains fixed over iterations.
    \item \textbf{Upper bound.} Simililar to the above, but greedily assign the upper bounds of the intervals to transitions while ensuring a probability mass less than or equal to one, until it is no longer possible. 
    The remaining probability mass is distributed uniformly over the remaining transitions.
    It remains fixed once initialized.
    \item \textbf{Random init.} At initialization, randomly select a single POMDP within the RPOMDP that remains fixed throughout the iterations.
    \item \textbf{Dom. Rand.} 
    At each iteration, randomly select a new POMDP within the RPOMDP, resembling \emph{domain randomization}~\cite{DBLP:conf/iros/TobinFRSZA17}. 
\end{description}
At each iteration, the POMDP used is described above.
The baselines and \rfscnet run for the same number of $50$ iterations.

\begin{table}[tbp]
	\centering
	\begin{tabular}{lrrrr}
\toprule
Instances    & \textit{Aircraft} & \textit{Avoid} & \textit{Evade} & \textit{Intercept} \\             \cmidrule(lr){1-5}
$|S|$        & $13552$           & $10225$        & $4261$         & $4802$             \\ \cmidrule(lr){1-5} %
$|Z|$        & $37$              & $6968$         & $2202$         & $2062$             \\ \cmidrule(lr){1-5} %
$|A|$ & $5$               & $4$            & $5$            & $4$              \\
\bottomrule
\end{tabular}

	\vspace{1em}
	\caption{Dimensions of each benchmark environment.}
	\vspace{2em}
	\label{tab:sizes}
\end{table}

\paragraph{Metric.}
We compare the \emph{robust values} $\mathcal{J}^{\fsc}_{\uTrsFun}$, of the FSCs $\fsc$ computed by the baselines, \ours, and \SCP. 
For \rfscnet and the baselines, we consider the best robust value found across the iterations.
As these methods include randomness in the sampling, initialization, and training, we report statistics of the robust value across $20$ seeds.
\SCP is not random given a fixed initialization of its hyperparameters, thus we only report a single value for its default settings.
\paragraph{Tools and Hyperparameters.}
We use the tools \storm~\cite{DBLP:journals/sttt/HenselJKQV22} for parsing the models and \prism~\cite{KNP11} to compute the RMDP values for the lower bounds in \Cref{tab:sota_comparison} and for robust value iteration on the robust Markov chain in robust policy evaluation.
We build and train the RNN and the QBN using {TensorFlow}~\cite{abadi2016tensorflow}.
The RNN cell is a gated recurrent unit (GRU)~\cite{DBLP:conf/emnlp/ChoMGBBSB14}.
For all the experiments, the simulation batch size is set to $I=256$, the maximum simulation length is set to $H=200$, and we run for a maximum of $50$ iterations. 
The RNN and QBN use an Adam optimizer~\cite{kingma2014adam} with a learning rate of $1\cdot 10^{-3}$.
The hidden size of the RNNs was set to $d=16$.
For solving the LP, we use the Gurobi solver~\cite{gurobi}.
The experiments ran inside a Docker container on a Debian 12 machine.
Our infrastructure includes an AMD Ryzen Threadripper PRO 5965WX machine with 512 GB of RAM.
We train the neural networks on the CPU.
The different seeds for the RNN-based methods were executed in parallel, each running on a single core.
Multi-threading in the Gurobi solver used by \SCP was enabled. 
In our initial tests, we considered hidden sizes $d\in\{3,16,64\}$, batch sizes $I \in \{128,256,512\}$, learning rates in the range of $[1\cdot 10^{-2}, 1\cdot 10^{-4}]$, and different number of iterations before arriving at our final values.
We used the same infrastructure and experimental setup across methods. %

\paragraph{Results.}
\Cref{fig:robust_evaluation} compares the performance of \rfscnet to the aforementioned baselines.
\Cref{tab:sota_comparison} shows \rfscnet's median and minimum performance when configured with a maximal memory size of $\memsize = 9$, compared to the \SCP method with two different sizes $\memsize\in\{3,9\}$.
The heatmaps in \Cref{fig:ex:heatmaps} showcase the effect of various memory sizes $\memsize$ on the performance of both \rfscnet and \SCP in \emph{Aircraft} and \emph{Evade}.
In these results, \rfscnet is equipped with \texttt{k-means++} clustering and \QMDP as supervision policy.
\Cref{fig:intercept:train:loss} shows the difference in training performance, in terms of the RNN and QBN losses, between \rfscnet and a baseline on \emph{Intercept}.
\Cref{fig:configurations} compares \rfscnet across multiple supervision policies and clustering configurations to a baseline on \emph{Intercept}.
In \ifappendix\Cref{app:expval}\else
Appendix F~\cite{DBLP:journals/corr/abs-2408-08770}\fi, we provide the complete set of results.
\subsection{Analysis}
We now address each research question based on the experiments.
\paragraph{\ref{q:robustness} Comparison to baselines.}
As seen in \Cref{fig:robust_evaluation}, \rfscnet outperforms all baselines on \emph{Aircraft} and \emph{Intercept}, reaching a median value that is lower than the first quartile of all baselines.
On average, {\rfscnet incurs lower expected costs than the baselines}.
\Cref{fig:intercept:train:loss} shows that in \emph{Intercept}, training on \adverb POMDPs, instead of a single fixed POMDP, is a more difficult learning target. 
Even though the training task is more challenging, \rfscnet still performs well.
The results are more ambiguous in the \emph{Evade} environment. 
The baselines can perform better, demonstrating that ignoring the model uncertainty may suffice in this environment.
Nonetheless, we still observe that {\rfscnet achieves at least the same robust performance as the baselines}.
On \emph{Avoid}, \rfscnet performs slightly worse than the baseline that is trained on the upper bound of the uncertainty set, while the remaining baselines clearly perform very poorly.
We conjecture that, by coincidence, the upper bounds provide an adequate approximation of the worst-case probabilities.
Without this particular initialization, \rfscnet still achieves comparable performance to this baseline.
Good baseline performance is not guaranteed, and the baselines may find much worse performing policies, as evidenced by the results obtained when trained on the lower bound or through domain randomization.
Thus, the baselines are unreliable as they are sensitive to the POMDPs used throughout training, while \rfscnet performs reliably across environments.

\begin{table}[tb]
        \resizebox{0.99\columnwidth}{!}{
        \begin{tabular}{@{}lllllll@{}}
\multicolumn{2}{@{Robust Values $\mathcal{J}^{\fsc}_{\uTrsFun}$}l}{} & \textit{Aircraft} & \textit{Avoid} & \textit{Evade} & \textit{Intercept} \\ 
\toprule
Lower Bound & (RMDP) &
$94.24$ & $18.05$ & $31.19$ & $16.99$  \\ 
\midrule
\multirow{2}{*}{\SCP} & \multicolumn{1}{l}{($\memsize=3$)} & $116.03$ & $20.07$ & $37.97$ & ${\textcolor{green!60!black}{\textbf{31.57}}}$ \\  %
 &  \multicolumn{1}{l}{($\memsize=9$)} & $116.58$ & $29.51$ & $39.78$ & $101.12$ \\ 
 \midrule
\multirow{3}{*}{\rfscnet} & (med.) & \textcolor{green!60!black}{$\textbf{103.91}$} & $\textcolor{green!60!black}{\textbf{18.29}}$ & $
\textcolor{green!60!black}
{\textbf{37.95}}$ & $47.82$ \\  %
 & (min.) & ${102.10}$ & ${18.20}$ & ${37.10}$ & $31.61$ \\
  & (iqr.) & $\pm 2.69$ & $\pm 0.12$ & $\pm 1.24$ & $\pm 26.97$ \\
 \bottomrule\\
\end{tabular}

        }
    \caption{
    The median (med.), minimum (min.), and 
    interquartile range (iqr.)
    of the robust values of 
    \rfscnet across $20$ seeds compared to those values of \SCP (lower is better). 
    RMDP denotes a lower bound on the robust value by computing the robust value of the underlying RMDP, assuming full observability.
    We highlight the best value between \SCP and the med. of \rfscnet.
    }
    \label{tab:sota_comparison}
    \vspace{2em}
\end{table}

\begin{figure}[tb]
    \centering
    \includegraphics[width=0.495\linewidth]{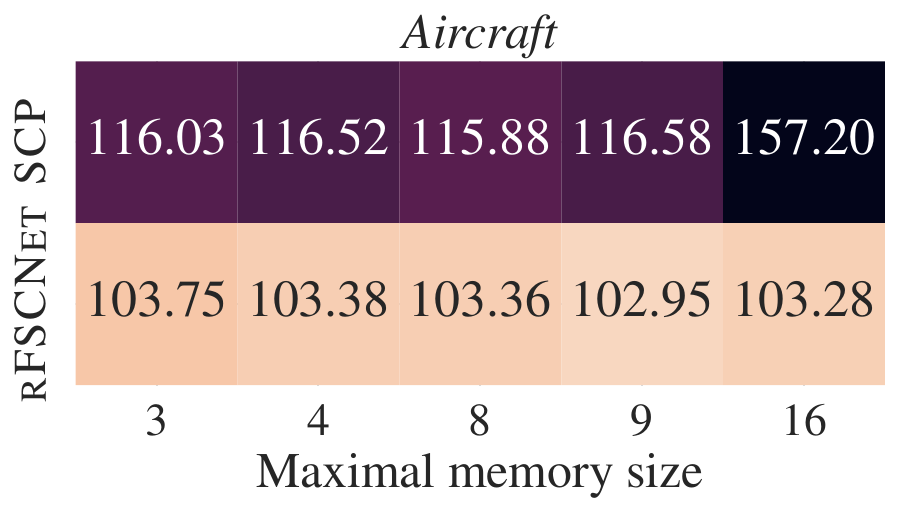}
    \includegraphics[width=0.495\linewidth]{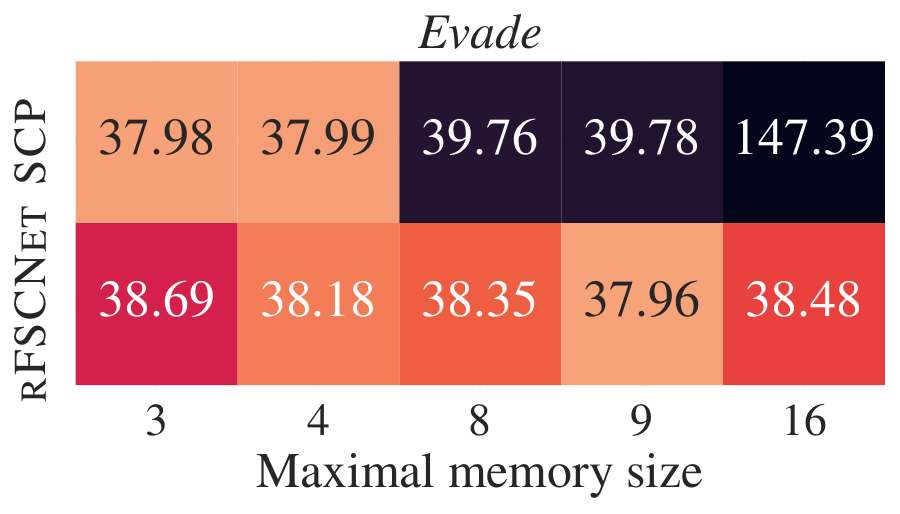}
    \caption{Heatmaps comparing the robust values $\mathcal{J}^{\fsc}_{\uTrsFun}$ under various memory~sizes $\memsize$ of \rfscnet (median across $20$ seeds reported) to \SCP's.
    Brighter colors (lower values) are better.
    With high memory sizes $\memsize$, \SCP performs worse, whereas \ours performs consistently across all sizes.
    }
    \label{fig:ex:heatmaps}
    \vspace{3em}
\end{figure}

\paragraph{\ref{q:comparison} Comparison with the state-of-the-art.}
As seen in \Cref{tab:sota_comparison}, in comparison to \rfscnet, \SCP performs comparably on \emph{Evade} and best on \emph{Intercept} when $\memsize = 3$. 
These results showcase that \SCP performs well if the memory is cherry-picked for the problems. 
In the case that the memory size in \SCP is set to $\memsize=9$, which is the same maximal memory size we set for \rfscnet in these results, \rfscnet significantly outperforms \SCP across all benchmarks.
Furthermore, \rfscnet significantly outperforms \SCP  on \emph{Aircraft} and \emph{Avoid} across both memory configurations.
Therefore, we conclude that {\rfscnet improves over the state-of-the-art} in these cases.

\begin{figure}[t]
  \centering
    \includegraphics[width=.98\linewidth]{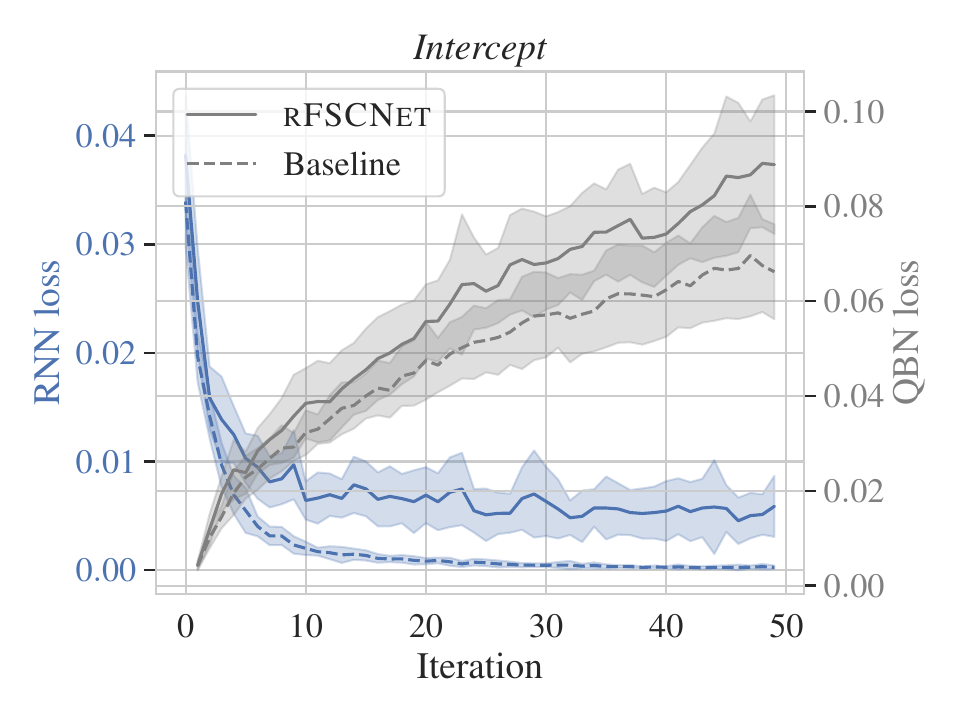}
        \caption{Comparing the \emph{post hoc} QBN and RNN training losses (averaged over the seeds) between a baseline running on a fixed POMDP and \ours on the RPOMDP on \emph{Intercept}. 
        Both are equipped with \QMDP and a QBN.
        }
        \label{fig:intercept:train:loss}
    \vspace{3em}
\end{figure}

\paragraph{\ref{q:memory} Memory size sensitivity.}
\Cref{tab:sota_comparison} 
indicates that \SCP performs worse with more memory, especially on \emph{Avoid} and \emph{Intercept}.
To investigate further, we conduct additional experiments to compare the performance of \SCP and \ours on \emph{Aircraft} and \emph{Evade} with increasing memory sizes.
The heatmaps in \Cref{fig:ex:heatmaps} show that \SCP also performs much worse on \emph{Aircraft} and \emph{Evade} when more memory is specified
A larger memory size implies more optimization variables for \SCP, which may be why its performance deteriorates, as \SCP may get stuck in worse local optima when we allow for more memory.
In contrast, {\rfscnet is not sensitive to specifying more memory nodes than necessary, exhibiting relatively consistent performance across the memory sizes}.
These results demonstrate the benefit of learning the memory structure instead of specifying it beforehand, as done in \SCP.

\begin{figure}[t]
\centering
     \includegraphics[width=.98\linewidth]{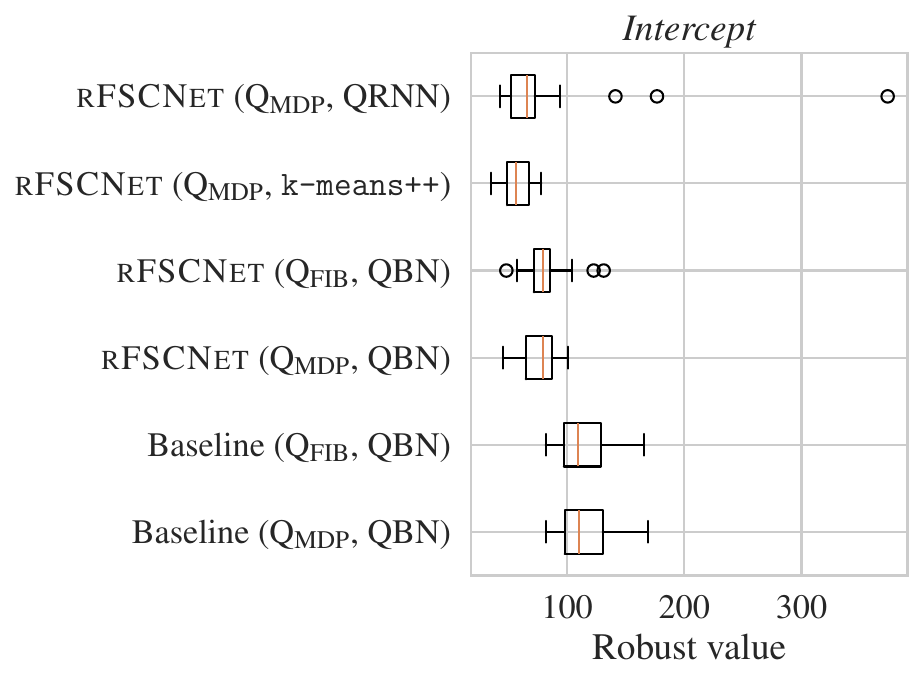}
     \caption{Boxplots of the best robust values across seeds on multiple configurations tested for \rfscnet on the Intercept RPOMDP and a baseline on a fixed (nominal) POMDP instance.
     }
     \label{fig:configurations}
     \vspace{3em}
\end{figure}

\paragraph{\ref{q:configuration} Configuration sensitivity.}
\Cref{fig:configurations} depicts the performance of \rfscnet 
across various configurations on \emph{Intercept}. 
\ifappendix\Cref{fig:app:conf} in \Cref{app:subsec:configurations}\else
Figure 6 in Appendix F.1~\cite{DBLP:journals/corr/abs-2408-08770}\fi\xspace shows these results also for the other three environments.
From our results, we did not observe a major difference between using \QMDP and \QFIB as supervision policies.
While sometimes producing better results, training the QBN \emph{end-to-end} proves less stable than \emph{post hoc}.
Overall, clustering with \texttt{k-means++} produces the best results in our benchmarks.
The results demonstrate that, while the configuration does impact performance, {\rfscnet 
performs consistently across configurations.}

\subsection{Discussion}

The performance of the FSCs is impacted by the quality of discretization of the hidden states of the RNN, and faulty extraction of the FSC from the RNN leads to finding a robust value that is not informative and, consequently, a pessimistic POMDP that is not helpful.
How to optimally extract finite-state representations of RNNs is still an open problem.
In this paper, we tested multiple options based on clustering. 
We emphasize that the \framework framework is modular and allows other methods that compute FSCs for POMDPs to be used instead.

Finally, we note that the performance of \rfscnet is limited by the quality of the supervision policies we compute during training.
This limitation could, for instance, explain why \rfscnet performs worst on \emph{Intercept}, as this benchmark relies on information gathering, an aspect on which \QMDP is known to perform poorly.
Nonetheless, \rfscnet's modularity allows for any POMDP policy to be applied as a supervision policy in the RNN's training procedure, allowing for trade-offs between quality and computational efficiency.

\section{Related Work}\label{sec:related:work}
Early works on RPOMDPs extended value iteration and point-based methods to account for the additional uncertainty~\cite{DBLP:journals/ai/ItohN07,DBLP:conf/icml/Osogami15}, or use sampling over the uncertainty sets~\cite{DBLP:conf/icra/BurnsB07a}.
\citet{DBLP:journals/siamjo/NakaoJS21} extend value iteration for \emph{distributionally robust} POMDPs, where the agent receives side information after a decision period, which is a less conservative setting.
Extensions to value iteration for RPOMDPs typically do not scale well to the large state spaces (up to $13000+$) we consider in this paper.
\citet{DBLP:conf/aips/NiL08,DBLP:journals/soco/NiL13} introduce a policy iteration algorithm for optimistic policies, which does not extend to the robust setting we consider. 
\citet{DBLP:conf/cdc/ChamieM18} consider robustifying a given POMDP policy to combat sensor imprecision.
Recent methods compute FSCs for finite sets of POMDPs through subgradients~\citep{hmpomdps} and for RPOMDPs via 
quadratic~\cite{DBLP:conf/ijcai/Suilen0CT20}
or sequential~\cite{DBLP:conf/aaai/Cubuktepe0JMST21}
convex programming, with the latter outperforming the former.
In contrast to our work, the convex optimization methods compute FSCs of a predefined size and structure and cannot handle the optimistic case.
RNNs have previously been used in a planning setting to compute FSCs for (non-robust) POMDPs~\citep{DBLP:journals/jair/Carr0T21}. 
Yet, robustness against model uncertainty was not considered, and, to the best of our knowledge, no method exists that exploits the learning capabilities of RNNs in a robust planning setting.

\section{Conclusion}\label{sec:conclusion}
In this paper, we presented \framework, a novel planning framework for RPOMDPs, and our algorithm, \ours, that is based on the PIP framework.
\ours utilizes RNNs to compute FSCs for the \adverb POMDPs selected by \framework, allowing the memory structure to be learned from data.
Our experiments show that our approach 
yields more robust policies than the baseline approaches.
Additionally, \rfscnet is less sensitive to over-parameterization of the memory size than \SCP, a state-of-the-art solver.
Furthermore, \rfscnet outperforms \SCP in a subset of the benchmarks considered in this paper.
Future work may investigate alleviating the limitation of the supervision policies by optimizing the RNN with a more sophisticated training objective or by considering more advanced supervision policies.

\begin{ack}
We thank the anonymous reviewers for their valuable feedback, Eline M. Bovy and Merlijn Krale for helpful discussions on earlier versions of this work, and Niels Neerhoff for providing the code from his master's thesis. 
This research has been partially funded by the NWO grant
NWA.1160.18.238 (PrimaVera), the FWO grant G0AH524N (SynthEx), and the ERC Starting Grant
101077178 (DEUCE).

\end{ack}

{%
\bibliography{literature}
}

\ifappendix
\appendix
\onecolumn

\section{Robust Markov Chain and Robust Policy Evaluation}\label{appx:rdp}
In this appendix, we provide an extended definition, including details on the computation of the robust value function $\Vmc$, and show that robust policy evaluation, by constructing and evaluating the policy on the robust Markov chain (\Cref{def:robust:markov:chain}), indeed provides a conservative upper bound.
Furthermore, we show that the static and dynamic uncertainty models coincide under our conditions, extending and building on results from\ifappendix~\citet{DBLP:journals/mor/Iyengar05} and~\citet{DBLP:phd/ndltd/Patek97}\fi.

\begin{definition}[Robust Markov chain and robust policy evaluation (extended)] \label{def:app:robust:policy:evaluation}
Given an RPOMDP $\cM = \braket{S,A,\uTrsFun, \RewFun, Z, \ObsFun}$ with initial state distribution (belief) $b_0$ and an FSC $\fsc = \braket{N,n_0, \memupdate, \actionmap}$, the robust state-values of $\cM$ under $\fsc$, $\underline{V}^{\fsc} \colon S\times N \to \mathbb{R}$, are given by the state-values in the \emph{robust Markov chain} $\mc = \braket{S \times N, b^{\fsc}_0,\uTrsFunMC, \RewFun^{\fsc}}$ where the state-space is the product of RPOMDP states~$S$ and FSC memory nodes~$N$, the initial state distribution is $b^{\fsc}_0(\braket{s,n}) = b_0(s)[n = n_0]$.
The uncertain transition function and cost functions are:
\begin{align*}
 \uTrsFunMC(\braket{s',n'} \given \braket{s, n}, a) &= \uTrsFun(s, a)(s') \fscjointprob(a, n' \mid s, n)\\
\uTrsFunMC(\braket{s',n'} \given \braket{s, n}) &=
 \sum_{a \in A} \uTrsFunMC(\braket{s',n'} \given \braket{s, n}, a)\\
 \RewFun^{\fsc}(\braket{s, n}) &=
 \sum_{a \in A} \actionmap(a \given n, \ObsFun(s))\RewFun(s,a),
\end{align*}
where addition and multiplication over intervals follow the standard rules for interval arithmetic\ifappendix~\cite{DBLP:journals/jacm/HickeyJE01}\fi.
For completeness, the uncertainty sets of the uncertain transition function of the RMC are defined as:
\begin{align}
    \uTrsFunMC(\braket{s,n}, a) &= \bigg\{\TrsFun(s, a) \actionmap(a \given n, \ObsFun(s)) 
    \in\Delta(S\times N) \;\big\vert\; \forall \braket{s',n'}\in S \times N \colon T(s'\mid s,a) \in \uTrsFun(s,a)(s') \wedge [n'=\memupdate(n, \ObsFun(s)] \bigg\}.\\
    \uTrsFunMC(\braket{s,n}) &= \bigg\{\sum_{a\in A}\TrsFun(s, a) \actionmap(a \given n, \ObsFun(s)) 
    \in\Delta(S\times N) \;\big\vert\; \forall \braket{s',n'}\in S \times N \colon T(s'\mid s,a) \in \uTrsFun(s,a)(s') \wedge [n'=\memupdate(n, \ObsFun(s)] \bigg\}.
\end{align}
Then, the robust state-values $\Vmc \colon S\times N \to \mathbb{R}_{\geq 0}$ are given by the robust Bellman equation and follows directly from robust dynamic programming on $\mc$:
\begin{align}
    \Vpes^{\fsc}(\braket{s,n}) &= 
    C^{\fsc}(\braket{s,n}) + 
    \underbrace{\sup_{\TrsFunMC(\braket{s,n})\in\uTrsFunMC(\braket{s,n})} \sum_{s' \in S}\sum_{n' \in N} \TrsFunMC(\braket{s',n'} \given \braket{s, n}) \Vpes^{\fsc}(\braket{s',n'})}_{\text{Inner problem}}.\label{eq:appx:robust:as:value}
\end{align}
\end{definition}

\begin{definition}[Weighted maximum norm\ifappendix~\cite{DBLP:journals/mor/BertsekasT91}\fi]
    A \emph{weighted maximum norm} for a vector $x \in \mathbb{R}^n$ is a norm of the form: $||x||^{w}_{\infty} = \max_{i\in\{1,\ldots,n\}} \nicefrac{|x_i|}{w_i}$, where $w \in \mathbb{R}^n$ is a non-negative real-valued vector. 
\end{definition}
In our case, we have that $||\Vmc||^{w}_{\infty} = \max_{\tuple{s,n} \in S \times N} \nicefrac{\Vmc(\tuple{s,n})}{w(\tuple{s,n})}$ where $w \colon S \times N \to \mathbb{R}_{\geq 0}$ is a non-negative real-valued function.

\begin{definition}
        A policy is called \emph{proper} if it reaches the target set $G$ with probability 1.
\end{definition}

\begin{lemma}
    Assume $\fsc$ is proper, then all nature policies on the RMC must be proper, since we assume graph reservation (intervals with lower bounds $> 0$).
    Furthermore, due to our assumptions, nature acts stationarily and has a finite action space, comprised of the extreme points of the convex polytopal uncertainty set at each $(\tuple{s,n}, a)$.
    Therefore, we can use a classic result and conclude that $\Vmc$ is a contraction with respect to the maximum weighted norm\ifappendix~\cite{DBLP:journals/mor/BertsekasT91,DBLP:phd/ndltd/Patek97,patek_stochastic_1999}\fi. 
    Thus, we have the following inequality for any two $\Vmc_1,\Vmc_2$, for all $\tuple{s,n} \in S\times N$:
    \[
    \Vmc_1(\tuple{s,n}) - \Vmc_2(\tuple{s,n}) \leq \beta ||\Vmc_1 - \Vmc_2||_{\infty}^{w}.
    \]
\end{lemma}
It may seem counterintuitive that the contraction still applies, as nature's objective is to maximize the agent's cost.
Note that nature can only change the probabilities, potentially slowing progress toward the goal and incurring higher costs. 
Still, it cannot alter the underlying graph of the problem.
This ensures that the policy remains proper and the expected number of steps to reach the goal decreases in expectation; therefore, the difference between the value function and the actual expected costs shrinks, which is the intuition behind the contraction.

When we compute $\Vpes^{\fsc}$, we use the dynamic uncertainty model, allowing nature to select different probabilities from the uncertainty sets at each iteration of dynamic programming. 
Combined with the rectangularity assumption over the product state-space of the RMC, it allows for efficient computation of the inner optimization problem by solving it via a \emph{bisection algorithm}\ifappendix~\cite[Section 7.2]{DBLP:journals/ior/NilimG05}\fi.
Note that after the convergence of dynamic programming on the RMC, the probabilities converge to a single value when the least fixed point is achieved\ifappendix~\cite{DBLP:journals/mor/Iyengar05,DBLP:journals/ipl/ChenHK13}\fi.
Below, we establish this formally.

\subsection{The Static and Dynamic Uncertainty Model Coincide}

\noindent
\textbf{Theorem 1. }
\emph{    In the robust Markov chain $\mc$, the value under the dynamic model $\dynV$ coincides with the value under a static model $\staV$.}

\begin{proof}

Let $\fsc = \fsctup$ be a proper FSC, \ie, reaches the set of goal states with prob 1. 
We must have, since we are minimizing and nature is maximizing, that $\staV \leq \dynV$.
Therefore, we must establish that $\staV \geq \dynV$.
Our proof uses a similar argument to the existing argument of\ifappendix~\citet{DBLP:journals/mor/Iyengar05}\fi\xspace for RMDPs, but extends it to the robust Markov chain constructed from an FSC policy for robust expected costs.

For notational simplicity, assume that $\actionmap \colon N \to A$ is deterministic yet still proper.
The following argument generalizes to our definition that uses $\actionmap \colon N \times Z\to \Distr(A)$.
Choose any $\epsilon > 0$ and $\pickT \colon S \times N \times A \to \Delta(S)$  with for all $\tuple{s, n} \in S\times N$ we have $\pickT(\tuple{s,n}, \actionmap(n)) \in \mathcal{T}(s, \actionmap(n))$, and the following inequality:

\begin{align}
    \label{eq:epsineq}
    \dynV(\tuple{s,n}) &\leq \bigg[C(\tuple{s,n}) +  \sum_{\tuple{s',n'}\in S\times N} \pickT(\tuple{s',n'}\mid \tuple{s,n}) 
    \dynV(\tuple{s',n'}) \bigg] + \epsilon,
\end{align}
using $\pickT(\tuple{s',n'}\mid \tuple{s,n}) = \pickT(s'\mid \tuple{s,n}, \actionmap(n)) 
    [n' = \memupdate(n, O(s))]$, and $C(\tuple{s,n}) = C(s, \actionmap(n))$.
The inequality states that the dynamic value has expected costs that is less than or equal to the expected dynamic value of the next state under $\pickT$ plus some $\epsilon > 0$.  
Let $\dynV_{\pickT}$ denote the non-robust value of $\fsc$ under $\pickT$:

\begin{align*}
    \dynV_{\pickT}(\tuple{s,n}) = \bigg( C(\tuple{s,n}) + \sum_{\tuple{s',n'}\in S\times N} \pickT(\tuple{s',n'}\mid \tuple{s,n}) \dynV_{\pickT}(\tuple{s',n'})\bigg)
\end{align*}

Note that $\dynV_{\pickT}$ is the non-robust expected cost, which follows from dynamic programming on the resulting (non-robust) Markov chain, which is a contraction under similar (but milder, since there are no actions) assumptions as $\dynV$.
The non-robust value (expected costs) under $\pickT$ must be less than or equal to the static value $ \dynV_{\pickT} \leq \staV$.
Thus, $\staV \geq \dynV$ follows if we show that $\dynV_{\pickT} \geq \dynV$.

Given \Cref{eq:epsineq}, we consider the following inequality:

\begin{align*}
    (\dynV - \dynV_{\pickT})(\tuple{s,n}) &\leq  \bigg(\bigg[C(\tuple{s,n}) +  \sum_{\tuple{s',n'}\in S\times N} \pickT(\tuple{s',n'}\mid \tuple{s,n})
    \dynV(\tuple{s',n'}) \bigg] + \epsilon\bigg)\\
    &- \bigg( C(\tuple{s,n}) +  \sum_{\tuple{s',n'}\in S\times N} \pickT(\tuple{s',n'}\mid \tuple{s,n})
    \dynV_{\pickT}(\tuple{s',n'})\bigg) \\
    &=  \bigg[\sum_{\tuple{s',n'}\in S\times N} 
    \pickT(\tuple{s',n'}\mid \tuple{s,n})
    \bigg( \dynV(\tuple{s',n'}) -\dynV_{\pickT}(\tuple{s',n'})\bigg) \bigg] + \epsilon.
\end{align*}

Iterating this bound for $n\in\N$ times and using the contraction properties of the value functions, we get:

\begin{align*}
    (\dynV - \dynV_{\pickT})(\tuple{s,n}) \leq \beta^n||\dynV-\dynV_{\pickT}||_\infty^{w} + \sum_{j=0}^{n-1} \beta^j \epsilon
\end{align*}

Since $\epsilon$ and $n$ are arbitrary, we may establish that $\dynV_{\pickT} \geq \dynV$ and therefore,  $\staV \geq \dynV$.
This concludes the proof, since we have that $\staV \geq \dynV$ and $\dynV\geq\staV$.
\end{proof}

Consequently, there exists a transition function $T_N \colon S \times N \times A\to \Distr(S)$ adversarially chosen by nature that induces the worst-case expected costs of the FSC $\fsc$.
This is an unusual transition function for a POMDP, as it depends on the memory update function of the FSC.
Therefore, in the main body of the paper, we introduce a method to find a $T$ that is still pessimistic but independent of $N$.

\subsection{On the soundness of the robust value}

Under strict $(s,a)$-rectangularity of the RPOMDP, we would have the following robust Bellman equation to solve:
\begin{align}
\Vsa(\braket{s,n}) = \RewFun^{\fsc}(\braket{s, n}) + \sum_{a\in A} \underbrace{\sup_{T(s,a)\in\mathcal{T}(s,a)} \Big\{\notag \sum_{s' \in S}\sum_{n' \in N}
 \TrsFunMC_T(\braket{s',n'} \given \braket{s, n}, a)
 \Vsa(\braket{s',n'}) \Big\}}_{\text{Inner problem}}.\label{eq:robust:bellman:sa}
\end{align}
The inner optimization problem is convex under $(s,a)$-rectangular and interval uncertainty sets, but solving it at each dynamic programming step requires $|S||A|$ linear programs.
Furthermore, it is unclear how to constrain the inner supremum so that the same probabilities are picked at each memory node $n\in N$ of the FSC while remaining tractable.
For computational tractability, we instead opt to assume full $(s,a)$-rectangularity on the product state-space of the RMC, effectively leading to $(\braket{s,n}, a)$-rectangularity.
Below, we establish that this provides a conservative upper bound.

For an initial belief $b_0$ and initial memory node $n_0$, both state-based value functions can be extended to the value in the initial belief:
\begin{align*}
    & \Vsa(\braket{b_0,n_0}) = \sum_{s\in S} b_0(s)\Vsa(\braket{s, n_0}), \\
    & \Vmc(\braket{b_0,n_0}) = \sum_{s\in S} b_0(s)\Vmc(\braket{s, n_0}).
\end{align*}

\noindent
\textbf{Theorem 2. }
\emph{
    Given the FSC $\fsc$, the robust state-values $\Vmc$ of the robust Markov chain $\braket{S \times N, b_0^{\fsc}, \uTrsFunMC, \RewFun^{\fsc}}$ provide a (conservative) upper bound on the value $\Vsa$ of $\fsc$ under $(s,a)$-rectangularity in the RPOMDP with initial belief $b_0$.
    That is, $\Vsa(\braket{s,n}) \leq \Vmc(\braket{s,n})$, and consequently $\Vsa(\braket{b_0,n_0}) \leq \Vmc(\braket{b_0,n_0})$.
    }

\begin{proof}
    We show that $\Vmc(\braket{s,n}) \geq \Vsa(\braket{s,n})$ for all $(\braket{s,n})\in S\times N$.
    Recall \Cref{eq:appx:robust:as:value}. Omitting the constant $C^{\fsc}(\braket{s,n})$, we rewrite the inner supremum of the equation as follows:
    \begin{align*}
    \Vmc(\braket{s,n}) &= C^{\fsc}(\braket{s,n}) + \sup_{\TrsFunMC(\braket{s,n})\in\uTrsFunMC(\braket{s,n})} \sum_{s' \in S}\sum_{n' \in N} \TrsFunMC(\braket{s',n'} \given \braket{s, n}) \Vpes^{\fsc}(\braket{s',n'})\\
    &= 
    C^{\fsc}(\braket{s,n}) + \sup_{\TrsFunMC(\braket{s,n})\in\uTrsFunMC(\braket{s,n})} \sum_{s' \in S}\sum_{n' \in N} \sum_{a \in A} 
    \TrsFunMC(\braket{s',n'}\mid\braket{s,n}, a)
    \Vpes^{\fsc}(\braket{s',n'})\\
    &= 
    C^{\fsc}(\braket{s,n})+\sup_{\TrsFunMC(\braket{s,n})\in\uTrsFunMC
    (\braket{s,n})} \sum_{a \in A} \sum_{s' \in S}\sum_{n' \in N} 
     \TrsFunMC(\braket{s',n'} \given \braket{s, n}, a)
    \Vpes^{\fsc}(\braket{s',n'}).
    \end{align*}
    Ignoring the constant $C^{\fsc}(\braket{s,n})$, under $(s,a)$-rectangularity of the RPOMDP, we can continue rewriting this supremum as:
    \begin{align*}
    & \sup_{\TrsFunMC(\braket{s,n})\in\uTrsFunMC(\braket{s,n})} \sum_{a \in A} \sum_{s' \in S}\sum_{n' \in N} 
    \TrsFunMC(\braket{s',n'} \given \braket{s, n}, a)
    \Vpes^{\fsc}(\braket{s',n'}) \\
    &=
    \sum_a\sup_{\TrsFunMC(\braket{s,n}, a)\in\uTrsFunMC(\braket{s,n}, a)} \sum_{s' \in S}\sum_{n' \in N} 
     \TrsFunMC(\braket{s',n'} \given \braket{s, n}, a)
    \Vpes^{\fsc}(\braket{s',n'})\\
    &\geq \left(\sum_a \sup_{T(s,a)\in\uTrsFun(s,a)} \sum_{s'\in S}\sum_{n' \in N} 
    \TrsFunMC_T(\braket{s',n'} \given \braket{s, n}, a)
    \Vpes^{\fsc}(\braket{s',n'}\right).\\
\end{align*}
Inserting the constant $C^{\fsc}(\braket{s,n})$ again, we derive:
\begin{align*}
    &C^{\fsc}(\braket{s,n})+\sum_a\sup_{T(s,a)\in\uTrsFun(s,a)} \sum_{s' \in S}\sum_{n' \in N} 
     \TrsFunMC_T(\braket{s',n'} \given \braket{s, n}, a)
    \Vpes^{\fsc}(\braket{s',n'}
    = \Vsa(\braket{s,n}).
\end{align*}
Since the inequality holds for each state-memory node pair, we also have for some initial belief $b_0$ and initial memory node $n_0$ that $\Vsa(\braket{b_0,n_0}) \leq \Vmc(\braket{b_0,n_0})$.
\end{proof}
Intuitively, the robust Markov chain, and thus its value function $\Vmc$, operates under $(\braket{s,n},a)$-rectangularity, meaning nature may choose a probability distribution for each state $s \in S$, memory node $n \in N$, and action $a \in A$ independently.
In the RPOMDP, and thus the associated value function $\Vsa$, nature operates under $(s,a)$-rectangularity, meaning it chooses probability distributions independently of the state $s$ and action $a$ but is restricted to choose the same probability distribution for each memory node $n$.
The latter is more restrictive to nature, hence nature has fewer options to adversarially play against the agent.
As a result, the agent's cost may be lower than when nature's choices depend on the agent's memory.
This difference in semantics may also be explicitly encoded in a partially observable stochastic game by making the agent's memory either observable or unobservable to nature\ifappendix~\cite{ijcai2024p740}\fi.

\section{On the deterministic observation function}
\label{app:detobs}
Similarly to what has been established previously for POMDPs in\ifappendix~\citet{ChatterjeeCGK16}\fi\xspace and for RPOMDPs in \ifappendix~\citet{ijcai2024p740}\fi, we morph an RPOMDP with uncertain stochastic observation into an equivalent RPOMDP with deterministic observations. 
Let $\mathcal{O} \colon S\times A\times S\to (Z \to \mathbb{I} \cup \{0\})$ be an uncertain observation function, and $\cM_o = \braket{S, A, \uTrsFun, \RewFun, Z, \uObsFun}$ its RPOMDP tuple.
From $\cM_o$ we now construct an RPOMDP $\cM^+ = \braket{S^+, A, \uTrsFun^+, \RewFun^+, Z, O^+}$, that is equivalent but polynomially larger in the size of the state space, where:
\begin{itemize}
    \item $S^+ = S \times Z$ is the state space,
    \item $\uTrsFun^+ \colon S^+ \times A \to (S^+ \to \mathbb{I}\cup\{0\})$, with probabilities $\uTrsFun^+(\tuple{s',z'}\mid \tuple{s,z}, a) = \uTrsFun(s'\mid s,a)\uObsFun(z'\mid s',a,s)$.
    \item $\RewFun^+ \colon S^+ \times A \to \mathbb{R}_{\geq 0}$ with $\RewFun(\tuple{s,z}, a) = \RewFun(s,a)$, and,
    \item $O^+ \colon S^+ \to Z$ with $O^+(\tuple{s,z}) = z$.
\end{itemize}
The multiplication of intervals follows the standard arithmetic rules\ifappendix~\citep{DBLP:journals/jacm/HickeyJE01}\fi.
Therefore, the assumption of a deterministic observation function in the main body of the paper is without loss of generality.

\section{Supervision Policies}
\label{app:supervision_policies}
This section elaborates on the POMDP approximations used for computing the supervision policies.

\paragraph{QMDP.}
The \QMDP algorithm\ifappendix~\cite{Littman1995}\fi\xspace is an effective method to transform an optimal MDP policy to a POMDP policy by weighting the (optimal) action values $Q^*$ of the MDP to the current belief~$b\in\beliefs$ in the POMDP $M\in\cM$:
\[
    Q_\text{MDP}(b, a) = \sum_{s\in S} b(s) Q^*(s,a) = \sum_{s\in S} b(s) \left( \RewFun(s,a) + \sum_{s'\in S} T(s'\mid s,a) V^{*}_{\text{MDP}}(s') \right),
\]
where $V^{*}_{\text{MDP}}$ is the optimal value of the MDP underlying the POMDP $M$. 

\paragraph{Fast-informed bound.}
The fast-informed bound (FIB)\ifappendix~\cite{hauskrecht2000value}\fi\xspace approximates the optimal value of the POMDP. It is tighter than the one given by \QMDP since it includes a sum over the observation of the next state.
The $Q$ values of FIB are defined as:
\[
Q_{\text{FIB}}(b, a) = \sum_{s\in S}b(s)\alpha^a(s) = \sum_{s\in S}b(s) \left( \RewFun(s,a) + \sum_{z\in Z}
\min_{a'\in A} 
\sum_{s'\in S} T(s'\mid s,a) [z = O(s')] \alpha^{a'}(s')\right),
\]
where $\alpha^a \colon S \to \R$ for each $a \in A$ is a linear function, or \emph{alpha-vector}, updated via:
\[
\alpha_{i+1}^{a}(s) = \RewFun(s,a) +  \sum_{z\in Z}\min_{a'}\sum_{s\in S} T(s'\mid s,a) [z = O(s')] \alpha_i^{a'}(s').
\]

\section{Network Architectures}
In this section, we provide more details on the neural network architectures. 
Our \textit{post hoc} QBN approach largely follows \ifappendix\cite{DBLP:conf/iclr/KoulFG19} and \cite{DBLP:journals/jair/Carr0T21}\fi, apart from differences mentioned in \Cref{subsec:extract:fsc}, \ie, the extraction procedure of FSCs.
We used a batch size of $32$ for both networks during stochastic gradient descent. 

\subsection{QBN}\label{app:qbn}
Similar to prior work\ifappendix~\cite{DBLP:journals/jair/Carr0T21}\fi, we employ a quantized bottleneck network (QBN)\ifappendix~\cite{DBLP:conf/iclr/KoulFG19}\fi.
It consists of an encoder $E \colon \hat{\cH} \to [-1, 1]^l$ that maps the output of the RNN to a latent encoding with $\tanh$ activation, where $l$ is the latent encoding dimension.
The latent encoding is then quantized by a function $q \colon [-1, 1]^l \to \quantvals^{\bndim}$, where $\quantvals$ is the finite set of possible discrete values, for instance, $\quantvals = \{-1, 0, 1\}$ for three-level quantization.
The \emph{bottleneck dimension} $\bndim$ is the number of quantized neurons.
Lastly, there is a decoder $D \colon \quantvals^{\bndim} \to \hat{\cH}$ to reconstruct the input given the quantized encoding. 
The QBN represents a function $\cQ \colon \hat{\cH} \to \hat{\cH}$ where $\cQ(\hat{h}) = D(q(E(\hat{h})))$ for all $\hat{h}\in\hat{\cH}$. 
We train the QBN to minimize the reconstruction loss, \ie, mean-squared error, on the RNN's memory representations derived from the histories in $\cD$.
The finite set of memory nodes extracted is formed by the Cartesian product ${N}=\times_{\bndim} \quantvals$, and ${n}=q(E(\hat{h}))\in{N}$ is the discrete memory representation. 
Therefore, the extracted FSC's memory size $|{N}| = |\quantvals|^{\bndim}$ is directly controlled by $\bndim$ and the quantization level, \ie, size of the set~$\quantvals$. 
Note that the quantization level can be changed to be 2-level, \ie, with $B = \{-1, 1\}$ using the $\sign$ function as $q$, resulting in different controller sizes.

To ensure the encoder $E$ maps to $[-1, 1]$ we use $\tanh$ activation.
The gradient of this activation function is close to one around the zero input. 
Thus, for the $3$-level quantization, we use a version $\tanh_{flat}$ of the $\tanh$ function in the encoder that is flatter around the zero input to allow for easier learning of quantization level $0$, given by\ifappendix~\cite{DBLP:conf/iclr/KoulFG19}\fi:
\begin{equation*}
    \tanh_{flat}(x) = 1.5 \tanh(x) + 0.5 \tanh(-3x).
\end{equation*}
To allow the gradient to pass through the quantization layer, we employ a simple straight-trough estimator that treats the quantization as an identity function during back-propagation\ifappendix~\cite{DBLP:conf/icassp/LiuLWZY23}\fi.
The quantization activation function was provided by the Larq library\ifappendix~\cite{Geiger2020}\fi.
The encoder and decoder use a symmetrical architecture with $\tanh$ activation. 
The networks were quite small. The input and output sizes of the encoder and decoder were set to the hidden size $d$ of the RNN, with intermediate layers of sizes $8\cdot \bndim$ and $4\cdot \bndim$. 
\subsection{RNN}
\label{app:rnn}
We use a Gated Recurrent Unit (GRU, \ifappendix\cite{DBLP:conf/emnlp/ChoMGBBSB14}\fi) as the RNN architecture. 
Although there is no clear consensus between the Long Short-Term Memory (LSTM, \ifappendix\cite{DBLP:journals/neco/HochreiterS97}\fi) architecture and the GRU, the latter has fewer parameters than the LSTM but does have the ability to learn long-term dependencies due to the \emph{forget gate}. 
The forget gate is known to combat \emph{vanishing gradients} that occur through the variant of stochastic gradient descent employed for sequential models, known as backpropagation through time.
The inputs to the RNN were put through a learnable embedding layer. 
We trained the RNN policy using the method in \cref{subsec:rnn:training} with a categorical cross-entropy loss implemented in TensorFlow.
To prevent exploding gradients in the RNN, we use a norm-clipped gradient and orthogonal weight initialization\ifappendix~\cite{DBLP:journals/corr/SaxeMG13}\fi\xspace in the recurrent layer of the GRU, as recommended by\ifappendix~\cite{DBLP:conf/icml/NiES22}\fi.
For the policy head, we append two fully connected layers with size 32 and ReLU activation before the softmax layer mapping to the distribution over actions. 

\section{Benchmark Descriptions}\label{app:benchmark_desc}
In this section of the Appendix, we describe the benchmarks studied in the paper.
All environments are adapted with uncertain transition functions.
The grid-world environments model the probability of taking multiple steps instead of a single one for each possible moving action, to which we assign the interval $[0.1, 0.4]$.
In \emph{Aircraft}, we have two uncertainties: the probabilities of the pilot's responsiveness and of the adversary changing direction, both mapping to the same $[0.6, 0.8]$ interval. 
The dimensions of the benchmarks are given in \Cref{tab:sizes}.
We specify the dimensions of the grid-worlds to the same sizes as set in \ifappendix\cite{DBLP:conf/aaai/Carr0JT23}\fi.
\subsection{Aircraft Collision Avoidance}

We consider a discretized and model-uncertain version of the aircraft collision avoidance problem\ifappendix~\cite{collision}\fi\xspace as introduced in\ifappendix~\cite{DBLP:conf/aaai/Cubuktepe0JMST21}\fi. 

\paragraph{Aircraft.}
We follow the discretization procedure exactly and base our model on \ifappendix\cite{DBLP:conf/aaai/Cubuktepe0JMST21}\fi, but slightly adapt it for our expected cost formulation.
The objective is to minimize the expected cost, which models avoiding a collision with an intruder aircraft while taking into account partial observability (sensor errors) and uncertainty with respect to future paths. 
Crashes incur an additional cost of $100$ over the usual cost incurred of $1$ for each action. 
Furthermore, the only sink states are the goal states $G\subseteq S$.

\subsection{Grid-worlds}
We consider the grid-worlds introduced by\ifappendix~\cite{DBLP:conf/cav/JungesJS20}\fi\xspace but reformulate them as an expected cost (SSP) objective. 
All actions incur a cost of $c=1$, with an additional penalty of $c=100$ when in a bad state.
Once again, the only sink states of the RPOMDP are the goal states $G \subseteq S$.

\paragraph{Avoid.}
The \emph{Avoid} benchmark models a scenario where a moving agent must keep a distance from patrolling adversaries that move with uncertain speed. 
Additionally, its sensor yields partial information about the
position of the patrolling adversaries.
The agent may exploit knowledge over the predefined routes of the patrolling adversaries.

\paragraph{Evade.}
\emph{Evade} is a scenario where a robot needs to reach a destination and evade a faster adversary. 
The agent has a limited range of vision but may scan the whole grid instead of moving, incurring the same cost as moving. 
A certain safe area is only accessible by the agent. 

\paragraph{Intercept.}
\emph{Intercept} is the opposite of Evade because
the agent aims to meet another robot before it leaves the grid via one of two available exits.
Once the target robot has exited, the agent incurs an additional penalty of $c=100$ for each step before reaching a goal state.
On top of the view radius, the agent observes a corridor in the center of the grid.

\section{Extended Experimental Evaluation}\label{app:expval}

Below, the baseline is trained in the middle of the uncertainty set, a (nominal) POMDP instance $M \in \cM$, where the intervals of the uncertainty sets are resolved to a value in the middle of the interval $[i,j]$ given by $\nicefrac{i+j}{2}$, taking into account that transitions must sum to $1$.

\subsection{Configuration Study}\label{app:subsec:configurations}
Due to its modularity, \rfscnet allows for different configurations that may have an effect on its performance.
In \Cref{app:tab:all}, we collect median and minimum results across different configurations of \rfscnet. 
The combination of \QMDP and \texttt{k-means++} proves best, which is what is shown in the table of \Cref{tab:sota_comparison} in the paper.
QRNN, the method that uses a QBN trained \textit{end-to-end}, did not perform successfully on all environments.
This is due to instability during training, caused by updating the QBN's parameters with the gradients calculated from training the RNN, see \Cref{subsec:rnn:training}. 
However, by directly encoding the clustering of the QBN into the RNN architecture during training, we observe an improvement in the median and minimum performance on the two successful runs, \emph{Avoid} and \emph{Intercept}, over training the QBN \textit{post hoc}.
We also show results for the \baseline when trained with the two different supervision policies.
In these results, the baseline is trained on the (nominal) POMDP that resides in the middle of the uncertainty set.
The full results in the form of boxplots are depicted in \Cref{fig:app:conf}.

\begin{table}[tbp]
    \centering
    \begin{tabular}{@{}llllllll@{}}
 &  & \multicolumn{4}{c}{\rfscnet} & \multicolumn{2}{c}{\titlecap{\baseline}}  \\ \cmidrule(lr){3-6} \cmidrule(lr){7-8}
\multicolumn{2}{l}{} & QRNN & \texttt{k-means++} & QBN & QBN & QBN & QBN \\
\multicolumn{2}{l}{} & \QMDP & \QMDP & \QMDP & \QFIB & \QMDP & \QFIB \\\toprule
\multirow{2}{*}{\textit{Aircraft}} & med & $\times$ & $103.30$ & $\textbf{102.95}$ & $103.41$ & $105.91$ & $105.83$ \\
 & min & $\times$ & $102.03$ & $101.91$ & $\textbf{101.88}$ & $104.66$ & $104.60$ \\\midrule
\multirow{2}{*}{\textit{Avoid}} & med & $19.90$ & $\textbf{18.51}$ & $20.19$ & $19.43$ & $18.83$ & $18.70$ \\
 & min & $18.62$ & $\textbf{18.16}$ & $18.57$ & $18.53$ & $18.39$ & $18.35$ \\\midrule
\multirow{2}{*}{\textit{Evade}} & med & $\times$ & $37.61$ & $37.96$ & $38.20$ & $\textbf{36.64}$ & $36.67$ \\
 & min & $\times$ & $36.65$ & $36.98$ & $37.07$ & $\textbf{36.06}$ & $36.11$ \\\midrule
\multirow{2}{*}{\textit{Intercept}} & med & $66.00$ & $\textbf{47.82}$ & $79.71$ & $79.51$ & $110.47$ & $109.36$ \\
 & min & $42.92$ & $\textbf{31.61}$ & $45.37$ & $48.16$ & $82.30$ & $81.66$ \\ \bottomrule
\end{tabular}

    \vspace{1em}
    \caption{Evaluation across multiple configurations for \rfscnet and a baseline trained on the (nominal) POMDP that resides in the middle of the uncertainty set.
    The values represent median (med.) and minimum (min.) robust values from the best FSCs computed of each run across 20 seeds.
    QRNN represents training the QBN \textit{end-to-end}, see \Cref{subsec:extract:fsc}.
    $\times$ indicates a run failed.
    \textbf{Bold} indicates the best (med/min) performance for each environment, \ie, across the rows.
    }
    \label{app:tab:all}
\end{table}

\begin{figure}[tbp]
    \includegraphics[width=0.49\textwidth]{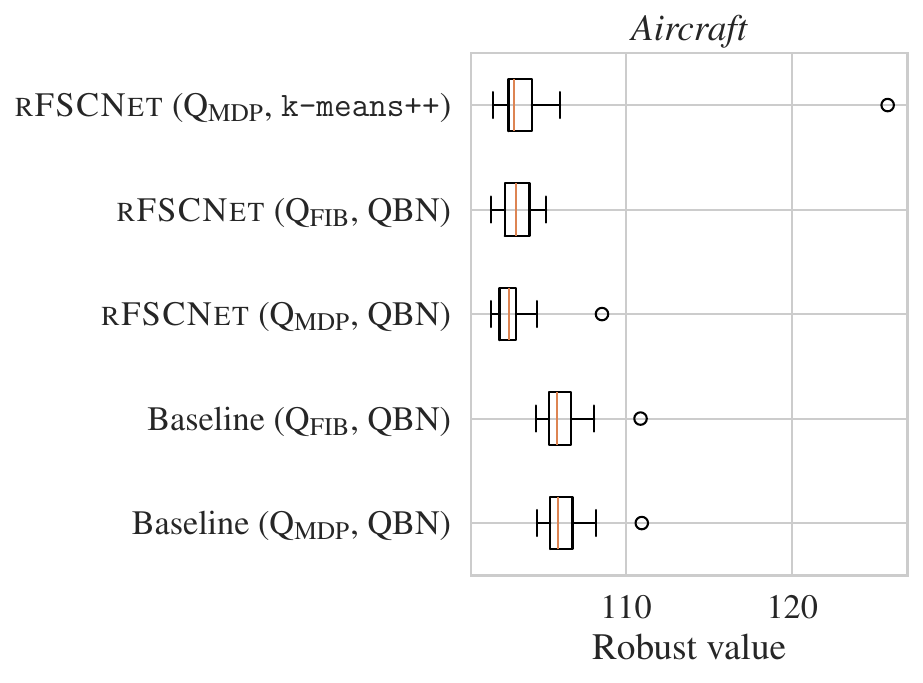}
    \includegraphics[width=0.49\textwidth]{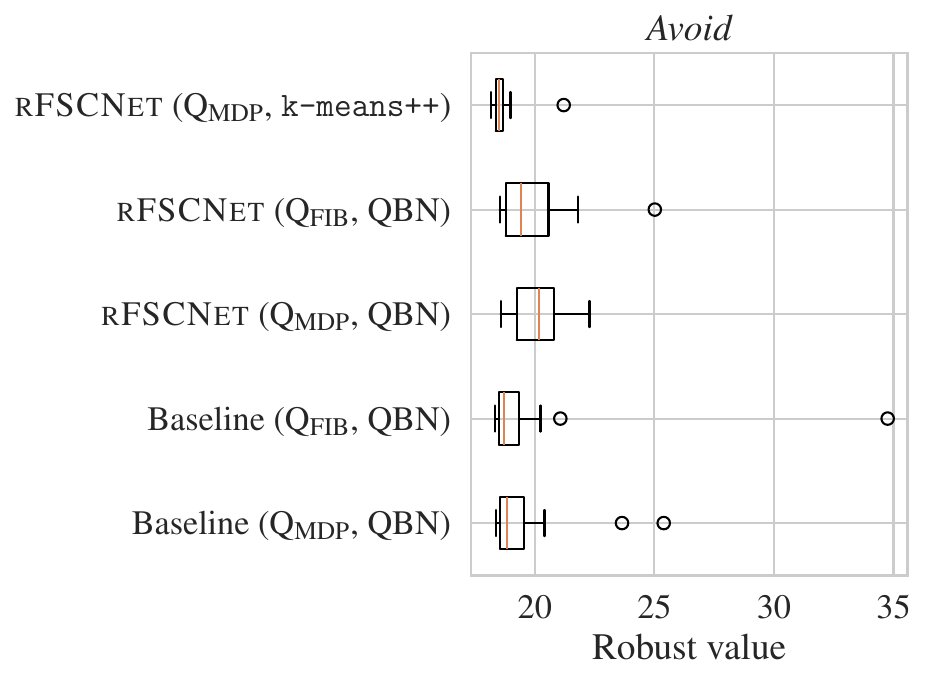}
    \includegraphics[width=0.49\textwidth]{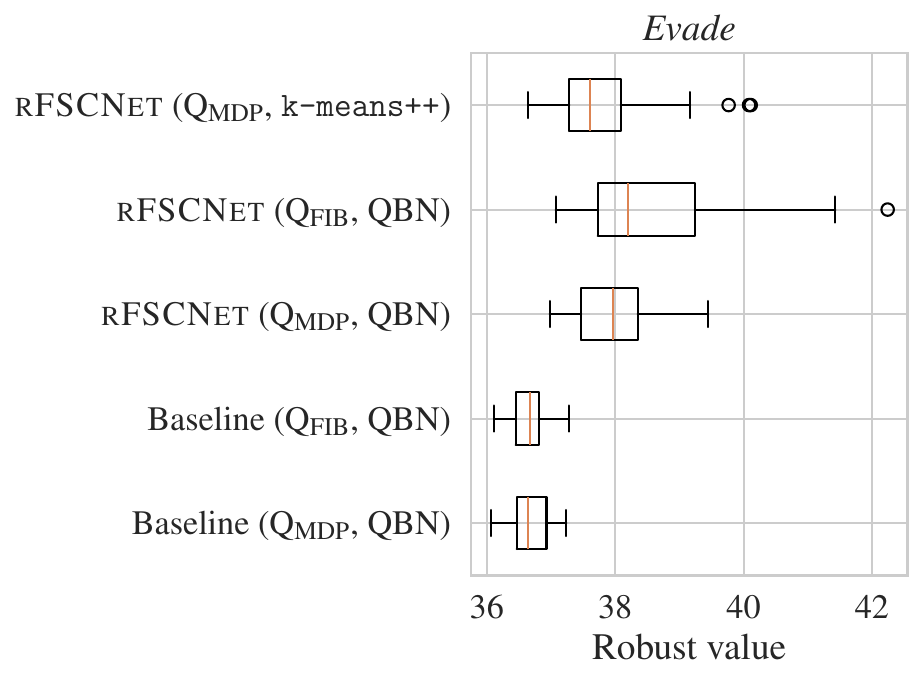}
    \includegraphics[width=0.49\textwidth]{plots_builder_output/all_comparison/with_qrnn/intercept-large.pdf}
    \vspace{1em}
    \caption{Comparison of the robust values between \rfscnet and a baseline trained across configurations. 
    For \textit{Avoid}, we plot without the run with QRNN, as it produces large outliers.
    }
    \label{fig:app:conf}
    \vspace{3em}
\end{figure}

\subsection{Extended Analysis on Various Memory Sizes}\label{app:memory}
In this subsection, we study the memory comparison between \rfscnet and \scp in more detail.
We chose \textit{Aircraft} and \textit{Evade}, as \scp appeared most consistent on these benchmarks.
In this study, \rfscnet ran with a \textit{post hoc} QBN and \QMDP as supervision policy.
We run for the maximal memory settings that we can restrict \rfscnet to when using the QBN, namely the values in the set $\memsize\in\{3,4,8,9,16\}$, which is the size of the sets that can be found through binary $\memsize\leq2^{\bndim}$ or ternary quantization $\memsize\leq3^{\bndim}$, see also \Cref{app:qbn}.
\Cref{fig:app:memory} extends the right-side plot in \Cref{tab:sota_comparison} with statistical details.
Namely, we plot the standard deviation around the median values in the heatmap and show the global min and max of each method.
We observe very stable performance for our method across the various memory sizes.
Both on \textit{Aircraft} and \textit{Evade}, \SCP shows relatively stable performance across memory sizes up to $\memsize=9$.
However, also on these benchmarks the performance drops when the required memory is set to a high level.
Evidently, \rfscnet does not suffer from the same phenomenon.
Furthermore, \rfscnet outperforms \textit{Aircraft} on all memory settings and performs similarly or better than \scp on \textit{Evade}.

\begin{figure}[tbp]
    \includegraphics[width=0.49\textwidth]{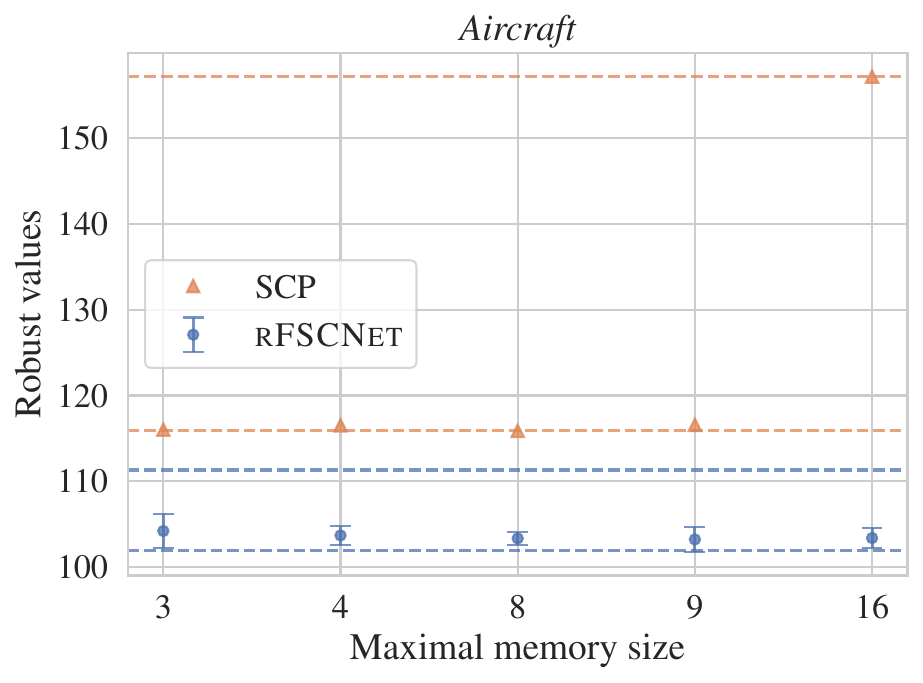}
    \includegraphics[width=0.49\textwidth]{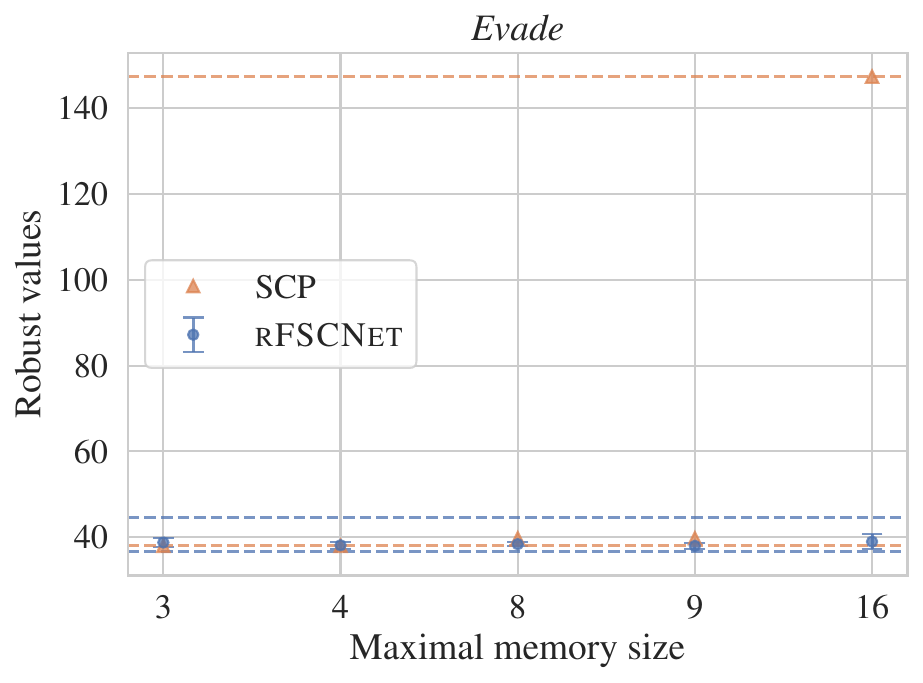}
    \caption{Comparison of the robust values between \rfscnet and \scp. 
    For \rfscnet, the error bars depict the standard deviation, and the dotted line shows for each method the global minimum and global maximum.
    }
    \label{fig:app:memory}
    \vspace{3em}
\end{figure}

\subsection{Loss Comparison}\label{app:loss_comparison}
\Cref{fig:app:losses} shows the RNN and QBN losses of the \baseline and \rfscnet on \textit{Aircraft} and \textit{Avoid}. Both runs employ a QBN and use FIB as the supervision policy.
The QBN is trained individually from the RNN, \ie, \textit{post hoc}.
The results show that as the RNN loss decreases, the QBN reconstruction loss increases.
This tells us that it gets increasingly hard to compress the hidden states of the RNN as they get more refined.
An intuition is that the RNN learns to use a larger part of $\hat{\cH}$ to represent the hidden states as training progresses, therefore making it harder to cluster the hidden states.
Alternatively, one could train the QBN \emph{end-to-end}.
However, as we elaborate in \Cref{app:subsec:configurations}, this approach suffers from instability during training and, therefore, did not successfully run on all environments.

\begin{figure}[tbp]
    \includegraphics[width=0.49\textwidth]{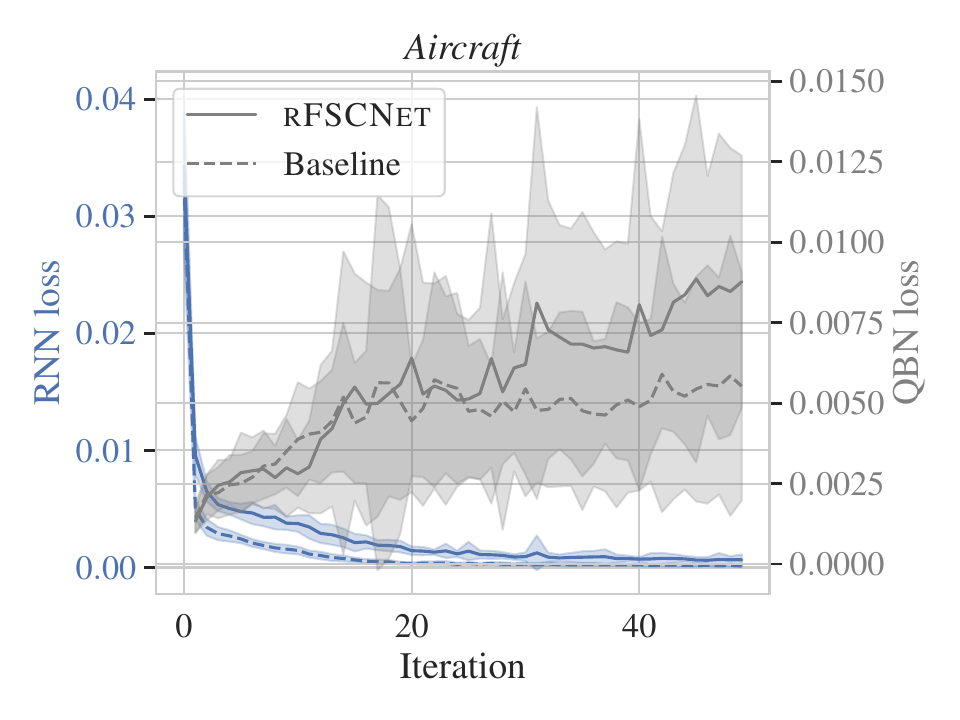}
    \includegraphics[width=0.49\textwidth]{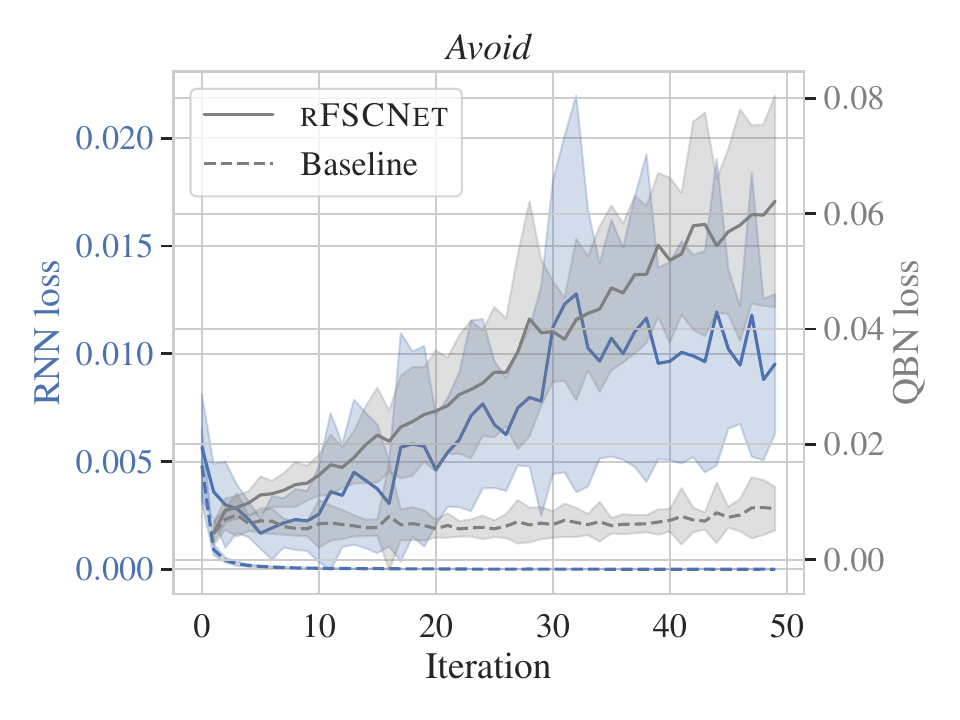}
    \includegraphics[width=0.49\textwidth]{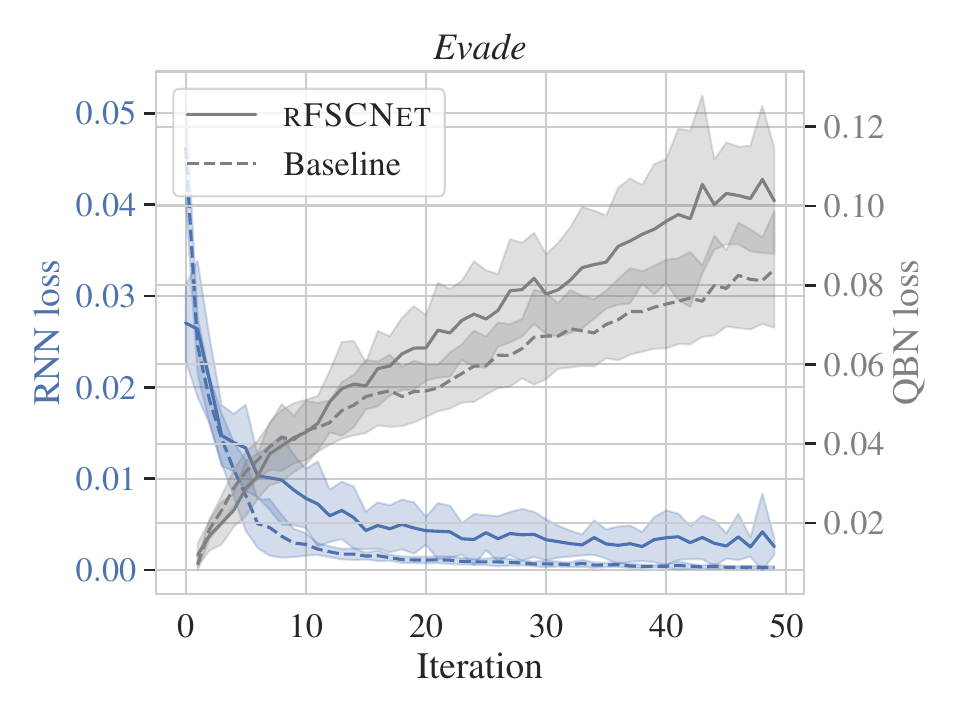}
    \includegraphics[width=0.49\textwidth]{plots_builder_output/loss_plots/intercept-large.pdf}
    \caption{Comparison of RNN and QBN losses between \rfscnet and a baseline over the iterations. 
    The line shows the mean over 20 seeds, and we plot the standard deviation around the mean for the RNN loss.
    On \textit{Aircraft} and \textit{Evade}, there is only a slight difference between the losses of the \baseline and \rfscnet.
    On the right, on \textit{Avoid}, a big difference is visible.
    }
    \label{fig:app:losses}
\end{figure}

\section{Run Times}\label{app:runtimes}
For completeness, we report the run times of each procedure for every environment in \Cref{tab:runtimes}. 
We report the run times using a QBN trained \textit{post hoc} and FIB, which is the most expensive configuration in terms of computations.
The RNN-based run times are averaged over the 20 seeds.
We would like to note that the times given for \SCP are \textit{user time} and do not account for the total \textit{CPU time} incurred by multi-threading. %
For the RNN-based approach, we see that the baseline is slightly faster in every environment except for \textit{Avoid}, as it does not execute Step 4 of our method from \Cref{subsec:lp}, and does not need to recompute the supervision policies as the POMDP is fixed. 
The run times naturally increase for larger FSCs because the Markov chain used for robust policy evaluation grows larger when the FSC has more memory nodes. 
Our method spends the majority of its time in its robust evaluation, executed by robust value iteration (robust dynamic programming) in PRISM.
Additionally, extraction from the QBN can take longer, as $|N|$ forward passes of the RNN are required.
Typically, the worse the policies found, the longer it takes to perform robust dynamic programming.
By comparing the run times of the baseline to our robust method, we see that the heuristic for finding worst-case instances does increase execution time.
Finally, we would like to point out that the run times for \SCP could be summed for a fair comparison, as running \SCP for only $\memsize=9$ yields much worse results than for $\memsize=3$.

\begin{table}[hbtp]
    \centering
    \caption{Average run times in seconds across $20$ runs for \rfscnet and the \baseline, and the \textit{user time} of the \SCP method on each environment. 
    Thus, both runtimes represent a form of user time.
    }
    \vspace{2em}
    {\begin{tabular}{@{}lrrrr@{}}
\toprule
                   & \rfscnet ($\memsize\leq 9$)    & Baseline ($\memsize\leq 9$)  & SCP ($\memsize=3$) & SCP ($\memsize=9$) \\ \midrule
\textit{Aircraft}  & $10562.51\pm156.03$  & $2518.39\pm158.76$  & $1133.8$ & $2169.3$ \\
\textit{Avoid}     & $9987.82\pm1209.82$ & $12778.36\pm1164.16$ & $2167.9$ & $6217.9$ \\
\textit{Evade}     & $5157.85\pm131.58$	   & $1281.85\pm58.73$   & $872.7$  & $3674.1$ \\
\textit{Intercept} & $2501.95\pm16.00$	   & $1624.33\pm12.49$   & $1884.9$ & $3243.5$ \\ \bottomrule
\end{tabular}
}
\label{tab:runtimes}
\end{table}

\fi

\end{document}